\PassOptionsToPackage{unicode}{hyperref}
\PassOptionsToPackage{hyphens}{url}
\documentclass[
  american,
  11pt,
  a4paper,
]{article}
\usepackage{xcolor}
\usepackage[margin=22mm]{geometry}
\usepackage{amsmath,amssymb}
\usepackage{iftex}
\ifPDFTeX
  \usepackage[T1]{fontenc}
  \usepackage[utf8]{inputenc}
  \usepackage{textcomp} 
\else 
  \usepackage{unicode-math} 
  \defaultfontfeatures{Scale=MatchLowercase}
  \defaultfontfeatures[\rmfamily]{Ligatures=TeX,Scale=1}
\fi
\usepackage{lmodern}
\ifPDFTeX\else
\fi
\IfFileExists{upquote.sty}{\usepackage{upquote}}{}
\IfFileExists{microtype.sty}{
  \usepackage[]{microtype}
  \UseMicrotypeSet[protrusion]{basicmath} 
}{}
\usepackage{setspace}
\makeatletter
\@ifundefined{KOMAClassName}{
  \IfFileExists{parskip.sty}{%
    \usepackage{parskip}
  }{
    \setlength{\parindent}{0pt}
    \setlength{\parskip}{6pt plus 2pt minus 1pt}}
}{
  \KOMAoptions{parskip=half}}
\makeatother
\usepackage{graphicx}
\makeatletter
\newsavebox\pandoc@box
\newcommand*\pandocbounded[1]{
  \sbox\pandoc@box{#1}%
  \Gscale@div\@tempa{\textheight}{\dimexpr\ht\pandoc@box+\dp\pandoc@box\relax}%
  \Gscale@div\@tempb{\linewidth}{\wd\pandoc@box}%
  \ifdim\@tempb\p@<\@tempa\p@\let\@tempa\@tempb\fi
  \ifdim\@tempa\p@<\p@\scalebox{\@tempa}{\usebox\pandoc@box}%
  \else\usebox{\pandoc@box}%
  \fi%
}
\def\fps@figure{htbp}
\makeatother
\ifLuaTeX
\usepackage[bidi=basic,shorthands=off]{babel}
\else
\usepackage[bidi=default,shorthands=off]{babel}
\fi
\ifPDFTeX
\else
\babelfont{rm}[UprightFont=*-regular.otf,ItalicFont=*-italic.otf,BoldFont=*-bold.otf,BoldItalicFont=*-bolditalic.otf]{texgyrepagella}
\fi
\ifLuaTeX
  \usepackage{selnolig} 
\fi
\usepackage{amsmath,amssymb,microtype}
\usepackage{float}
\usepackage{placeins}
\usepackage{caption}
\DeclareCaptionFont{integrated}{\fontsize{9}{10.6}\selectfont}
\usepackage{cite}
\usepackage{bookmark}
\IfFileExists{xurl.sty}{\usepackage{xurl}}{} 
\hypersetup{
  pdftitle={From Novel Combinations to Informative Absence},
  pdfauthor={Yanzhe Zhao},
  pdflang={en-US},
  hidelinks,
  pdfcreator={LaTeX via pandoc}}

\title{From Novel Combinations to Informative Absence}
\usepackage{etoolbox}
\makeatletter
\providecommand{\subtitle}[1]{
  \apptocmd{\@title}{\par {\large #1 \par}}{}{}
}
\makeatother
\subtitle{Structural permission, communicative opportunity, and form
choice in language learning}
\author{Yanzhe Zhao\\[2pt]
\small School of Future Technology, Tianjin University\\[1pt]
\small\href{mailto:justinzhaoyz@outlook.com}{justinzhaoyz@outlook.com} \enspace\textperiodcentered\enspace\href{https://orcid.org/0009-0003-5342-6611}{ORCID 0009-0003-5342-6611}}
\date{\small Author preprint}

\makeatletter
\patchcmd{\thebibliography}{\section*{\refname}}{}{}{}
\makeatother

\begin{document}
\maketitle

\setstretch{1.12}

\section{Abstract}\label{abstract}

Language learners can extend familiar grammatical patterns to expressions they have never heard, yet they also learn where such generalizations should stop. How does the continued absence of an expression become evidence that it is restricted? We propose a generative theory in which the value of an absence depends on the circumstances in which the expression could have been produced. We distinguish three components of language use: whether an expression is structurally permitted, whether the communicative situation provides an opportunity to express its meaning, and whether the speaker selects that particular form. In thirty artificial grammars, words that predict similar continuations support generalization to unseen word--construction combinations (mean AUC 0.945 after 120 sentences). We derive a joint evidence law for omissions across uncertain contexts and prove that, at fixed exposure, distributing omissions across conditionally independent contexts provides stronger evidence than concentrating them in one persistent context. Positive observations of related expressions can also raise expected use, making subsequent omissions more informative. As a result, a candidate with stronger initial structural support can become less plausible than another candidate if it is also expected to occur more frequently; in a fixed-rate example, their ordering reverses after six omissions. We further show that occurrence counts identify only the product of semantic-opportunity and form-choice probabilities, whereas recording opportunities separately identifies both conditional on permission. Reanalyses of artificial-language experiments with adults and children show that learners become more reluctant to generalize a familiar word when a competing construction expresses the same meaning. Across 399 participants, their judgments of restricted combinations also improve predictions of their own production choices, increasing the mean predictive log probability by 0.036 nats per production. In naturalistic speech from eleven families across six corpora, English article use varies systematically across recording sources, even after accounting for word identity, grammatical number, and syntactic role. Analyses of children's causative judgments in five languages further reveal how the predictive contributions of lexical frequency, semantics, and competition between forms vary across languages. Together, these findings connect the discovery of grammatical possibilities with the acquisition of their limits through a model of the circumstances in which an expression could have been produced.

\textbf{Keywords:} language acquisition; structural generalization;
indirect negative evidence; statistical preemption; context diversity;
hierarchical Bayesian inference.

\section{1. Introduction}\label{introduction}

Speakers can understand and produce expressions they have never encountered, yet they are also sensitive to limits on the grammatical patterns they generalize. For example, English speakers readily accept both \emph{She told him the story} and \emph{She told the story to him}. A learner might therefore extend the double-object pattern to another verb with a related meaning, such as \emph{explain}. In ordinary English, however, speakers strongly prefer \emph{She explained the story to him} over \emph{She explained him the story}. How does a learner discover such distinctions, especially when linguistic experience provides relatively little explicit correction? The ability to generalize productively while learning the limits of generalization is central to the problem often called Baker\textquotesingle s paradox \cite{art10}. In particular, a learner must determine when an unfamiliar combination is a genuine possibility and when its continued absence gives a reason to reject it.

One source of evidence about a new combination lies in the linguistic environments of its component words. Studies of distributional learning have shown how shared environments can reveal syntactic categories and relationships among forms \cite{art01,art11}. A word may never have occurred in a particular construction, but it may nevertheless predict the same subsequent linguistic material as words that are known to occur in that construction. We use a mechanism called Shared-Context Bridging (SCB) to formalize this idea: the separate histories of two words can provide evidence for their compatibility when those histories have informative predictive consequences in common. Such evidence helps a learner infer that a combination is structurally possible before observing it directly. A further question remains: if the combination is possible, how often should a learner expect to hear it?

Research on the restriction of linguistic generalizations suggests that this expectation depends partly on what speakers are trying to communicate \cite{art12,art13}. Consider the unfamiliar double-object use of \emph{explain}. Hearing the verb many times in any construction supplies evidence of its familiar uses, a possibility captured by accounts of entrenchment \cite{art18}. Hearing speakers repeatedly use an alternative construction to convey the meaning that the unfamiliar form would convey supplies a more specific kind of evidence, known as statistical preemption \cite{art14,art15,art16,art17}. Samara and colleagues \cite{art02} directly contrasted these conditions in artificial-language experiments and found stronger restriction under meaning-matched competition in adults and children. Research on sampling assumptions likewise shows that the significance of unobserved possibilities depends on how the learner believes the examples were generated \cite{art03,art04}. Hierarchical Bayesian accounts have further shown how learners can combine information about individual words with information about more abstract constructional patterns \cite{art10}. Together, these lines of work suggest that the learner must take into account both the distribution of the forms heard and the process that gives rise to those observations.

We propose that three distinct components of this process are relevant to interpreting an absence. The first is \textbf{structural permission}: does the grammar license the candidate word--construction combination? The second is \textbf{communicative opportunity}: does the situation provide a reason to convey the meaning that this expression would serve? The third is \textbf{form choice}: when that meaning is relevant and the expression is permitted, how likely is a speaker to select this particular form from the available alternatives? A learner who never hears the double-object use of \emph{explain}, for example, may face several possible explanations. The construction may be unavailable for this verb; situations calling for an explanation to another person may be uncommon; or speakers may generally prefer the prepositional form when describing such situations. All three can produce the same observable absence. Distinguishing among them requires expectations about how often relevant meanings arise and how speakers choose among forms. These expectations may differ across communicative contexts, while also reflecting tendencies shared across them.

This account connects positive generalization and restrictive learning. Evidence of structural compatibility can give an unfamiliar expression a high initial probability of being permitted. Separate evidence that relevant meanings and expressions occur frequently can raise the expected use of that candidate if it is permitted. If it then remains absent, each omission becomes more surprising. In some circumstances, the candidate that initially seemed more plausible can consequently become less plausible than another candidate with a lower expected usage rate. The force of such evidence also depends on how the observations are organized: repeated omissions in a single context can share a common explanation, whereas omissions across different contexts may constrain that explanation more strongly. We derive a joint mathematical law for these two forms of experience, an exact comparison at fixed exposure, and predictions about how positive observations elsewhere change the interpretation of later omissions. We also show why recording which meanings are available in a scene can separate learning about opportunities from learning about form choice.

We investigate the framework through a series of complementary computational and empirical studies. We begin with controlled artificial grammars, asking whether words that have never occurred together can be recognized as structurally compatible through the continuations they predict in common. We then reanalyze artificial-language experiments with adults and children to examine how competition between expressions conveying the same meaning influences restriction, and how learners' judgments of restricted combinations relate to their own production choices. The hierarchical model also predicts that the significance of an omission depends on how usage rates vary across communicative environments. We examine this premise in naturalistic child-directed and child speech by comparing article use across families and recording collections while accounting for lexical and grammatical properties.

The distinction between structural compatibility and expected usage raises a further question: what information helps learners estimate how frequently a particular expression will occur? We investigate this question by predicting article-use rates for previously held-out English nouns from their short linguistic contexts, and by comparing the contributions of lexical frequency, semantics, and competition to children's judgments of causative constructions across five languages. The mathematical and computational analyses derive quantitative predictions about the distribution of omissions across contexts, the influence of positive experience on later expectations, and changes in the relative plausibility of candidate expressions. The behavioral and naturalistic analyses establish empirical foundations for these mechanisms and guide experimental designs that bring them together.

Our central proposal is that learners interpret what they hear in relation to what they could reasonably have expected to hear. An absence acquires evidential force through the opportunities that might have generated the missing expression. By modeling the distribution, uncertainty, and observability of those opportunities, we connect the discovery of grammatical possibilities with the gradual acquisition of their limits.

\section{2. Theory}\label{theory}

\subsection{2.1 Structural permission, semantic opportunity, and form
choice}\label{structural-permission-semantic-opportunity-and-form-choice}

Consider a learner evaluating a particular word--construction combination. Before hearing the combination itself, the learner may already have evidence that it is grammatically possible. We represent the candidate\textquotesingle s permission by a binary variable \(Z\in \{0,1\}\), and the learner\textquotesingle s initial structural evidence by \(S\). The probability of permission after observing \(S\) is \(\pi (S)=P(Z=1\mid S)\). This probability expresses the learner's initial confidence that the combination is grammatically licensed.

Even a permitted expression will be heard only under certain circumstances. A communicative setting \(C_{t}\) may or may not provide an opportunity to express the relevant meaning. Let \(O_{t}\in \{0,1\}\) indicate whether such an opportunity is present on observation occasion \(t\), and \(Y_{t}\in \{0,1\}\) indicate whether the speaker produces the target form. We define

\[q_t=P(O_t=1\mid C_t,S),\qquad r_t=P(Y_t=1\mid O_t=1,Z=1,C_t,S).\]

Here \(q_{t}\) is the probability that the relevant meaning is available, and \(r_{t}\) is the probability of choosing the target form when that meaning is available and the target is licensed. We treat semantic opportunity as independent of the target\textquotesingle s permission conditional on the setting and structural evidence. In the basic model, a target occurrence requires both permission and an opportunity. Consequently, the probability of observing the target on an occasion, conditional on permission, is

\[P(Y_t=1\mid Z=1,C_t,S)=q_tr_t=\theta_t.\tag{1}\label{eq:1}\]

We refer to \(\theta _{t}\) as the effective target-use probability. An observation occasion is a sample from the communicative environment; whether it contains a relevant semantic opportunity may be unobserved. Thus the same omission can arise either because there was no reason to express the relevant meaning or because the speaker selected another form. Learning from omissions involves evaluating these possibilities as well as the possibility that the target is prohibited.

To quantify the evidence in a series of omissions, consider first the hypothesis that the target is permitted. Conditional on a particular path of usage probabilities, the omission on occasion \(t\) has probability \(1-\theta _{t}\). Under conditional independence of emissions given this path, a sequence \(D_{0}\) containing no target occurrences has likelihood

\[L(D_0\mid S)=\mathbb E\!\left[\prod_t(1-\theta_t)\mid Z=1,S\right].\]

The expectation averages over the usage-rate paths that the learner considers possible. Under the alternative hypothesis of strict prohibition, the same sequence has likelihood one. Bayes\textquotesingle{} rule therefore gives

\[\operatorname{logit}P(Z=1\mid D_0,S)=\operatorname{logit}\pi(S)+\log L(D_0\mid S).\tag{2}\label{eq:2}\]

An omission sequence lowers the log odds of permission by \(I=-\log L\), measured in nats (natural-log units). The less likely the observed omissions would have been under permission, the greater their evidential force. Equation~\eqref{eq:2} therefore connects two stages of learning: structural experience establishes the initial belief \(\pi(S)\), and subsequent observations update that belief through the omission likelihood \(L(D_0\mid S)\). The calculation of this likelihood depends on the learner's uncertainty about expected use, which we develop in Sections 2.3--2.4. Supplement S1 presents the full joint model and extends it to occasional production errors under prohibition. We begin by examining how the learner can obtain structural evidence for a combination that has never been encountered.

\subsection{2.2 Shared continuations supply structural support}\label{shared-continuations-supply-structural-support}

The preceding model assumes that learners have some evidence about whether a novel combination is structurally possible. One source of such evidence is the linguistic material that different words predict. Even when two words have never occurred together, their separate occurrences may be followed by similar continuations. Shared-Context Bridging (SCB) uses these common predictions to assess their compatibility.

To make this idea precise, consider artificial utterances containing a determiner \(D\), a noun \(N\), and an informative continuation \(U\). Some nouns occur only after a neutral initial frame, so that their direct pairing counts with any determiner are zero. Nevertheless, the learner can estimate the distributions of continuations associated with these nouns and with the determiners. Let \(\hat{p}_{U}(u)\) denote the overall frequency of continuation u. We compare the two conditional continuation distributions after subtracting this background frequency, weighting each common deviation by the inverse background frequency:

\[S(D,N)=\sum_u\frac{[\hat p(u\mid D)-\hat p_U(u)][\hat p(u\mid N)-\hat p_U(u)]}{\hat p_U(u)}.\tag{3}\label{eq:3}\]

The score is high when the determiner and noun tend to predict the same departures from the overall continuation distribution. In particular, it can distinguish words sharing category-linked continuations from words whose predictions differ, even when their direct pairing histories are equally empty.

To see how informative this cue can be, suppose there are \(K\) balanced latent categories and that a fraction \(\epsilon\) of continuations is generated independently of category. The category-specific part of each centered distribution has weight \(1-\epsilon\). In this setting, the population difference between within-category and between-category scores is

\[\Delta S=K(1-\epsilon)^2.\tag{4}\label{eq:4}\]

The separation decreases as continuations become less informative about category. Section 4.1 examines this relationship in simulated grammars, including a manipulation that disrupts shared predictions while preserving the observed word-pair structure. The resulting compatibility score provides structural evidence \(S\), from which a learner's initial permission probability \(\pi(S)\) can be specified through a suitable probabilistic mapping. Estimating how frequently a permitted expression will occur involves a second set of expectations, informed by the availability of relevant meanings and the speaker's choices among alternative forms. We now examine how uncertainty about these expectations affects the interpretation of omissions in Equation~\eqref{eq:2}.

\subsection{2.3 Persistent local uncertainty}\label{persistent-local-uncertainty}

A learner who knows exactly how often a permitted form should occur can evaluate omissions directly. For a known fixed effective rate \(\theta _{0}\), \(n\) independent omissions have likelihood \((1-\theta _{0})^n\), so their evidence against permission increases linearly with \(n\). The same likelihood holds if rates are independently renewed at every observation and have known mean \(\theta _{0}\).

In many communicative settings, however, the learner may be uncertain about how often a particular expression is used. Several observations from the same speaker or a persistent situation may share a local usage rate. If a target is repeatedly absent, the learner may increasingly attribute that absence to a low rate in that setting. To represent this possibility, suppose a single unknown effective rate \(\theta\) persists across observations within a context and \(\theta \sim \operatorname{Beta}(a,b)\), with \(a,b>0\). Integrating the omission likelihood over this rate yields

\[L_n=\frac{B(a,b+n)}{B(a,b)}\sim\frac{\Gamma(a+b)}{\Gamma(b)}n^{-a},\qquad I_n=a\log n+O(1).\tag{5}\label{eq:5}\]

After \(n\) omissions, the next-omission probability under permission is (b+n)/(a+b+n). As the omissions accumulate, higher local usage rates become less plausible and the learner increasingly expects the next observation to omit the target as well. Each additional omission therefore provides less new evidence against permission. In contrast to the linear accumulation of evidence at a known rate, uncertainty about a persistent local rate gives logarithmic accumulation. The next question is how this effect changes when the learner encounters several contexts whose usage tendencies may be related.

\subsection{2.4 A joint law for shared uncertainty across contexts}\label{a-joint-law-for-shared-uncertainty-across-contexts}

So far, the model has distinguished known usage rates from an uncertain rate that persists within a single context. A language learner, however, encounters many different settings. A speaker who rarely uses an expression may differ from another speaker, yet the two may also share conventions and communicative tendencies. Learning about the overall use of an expression can therefore change expectations about multiple contexts at once.

We capture these two levels of uncertainty by assigning each context \(c\) its own effective usage rate \(\theta _{c}\) and allowing the distribution of these rates to depend on an uncertain shared mean \(\mu\):

\[\mu\sim\mathrm{Beta}(A,B),\qquad\theta_c\mid\mu\sim\mathrm{Beta}(\kappa\mu,\kappa(1-\mu)),\quad A,B,\kappa>0.\tag{6}\label{eq:6}\]

The mean of a context-specific rate conditional on \(\mu\) is \(\mu\). The parameters A and \(B\) determine prior beliefs about this shared mean, and \(\kappa\) controls how concentrated the local rates are around it (larger \(\kappa\) gives less local variation at fixed \(\mu\)). Rates in distinct contexts are independent conditional on \(\mu\), while observations within a context share the same local rate. This hierarchy allows the learner to entertain both the possibility that an individual context has low use and the possibility that low use is common across contexts.

Suppose there are \(C\) contexts, each containing \(n\) target omissions. Given \(\mu\), the probability of \(n\) omissions in one context is the Beta-averaged factor \(m_{n}(\mu )\). Conditional independence allows us to multiply these factors across contexts before averaging over the shared mean. Define

\[m_n(\mu)=\prod_{j=0}^{n-1}\left(1-\frac{\kappa\mu}{\kappa+j}\right),\qquad L_{C,n}=\mathbb E_\mu[m_n(\mu)^C],\]

\[d_n=\sum_{j=0}^{n-1}\frac{\kappa}{\kappa+j}=\kappa[\psi(\kappa+n)-\psi(\kappa)].\tag{7}\label{eq:7}\]

Here \(\psi(x)=\mathrm{d}\log\Gamma(x)/\mathrm{d}x\) is the digamma function. The quantity \(d_n\) sums a sequence of decreasing weights associated with repeated omissions within a context. The first observation contributes a weight of one, and successive observations contribute progressively smaller weights. Mathematically, \(d_n\) is the coefficient of \(\mu\) in the first-order expansion of \(-\log m_n(\mu)\) near \(\mu=0\). It therefore characterizes the accumulation of evidence in the low-usage region that becomes increasingly important during prolonged absence. This leads to a natural question: how does encountering additional contexts compare with observing more omissions within a familiar context?

\textbf{Proposition 1 (joint evidence law).} Fix positive A, \(B\), \(\kappa\). For any integer sequences \(C,n\geq 1\) with \(Cd_{n}\to \infty\),

\[\boxed{L_{C,n}\sim\frac{\Gamma(A+B)}{\Gamma(B)}(Cd_n)^{-A},\qquad I_{C,n}=A\log(Cd_n)+O(1).}\tag{8}\label{eq:8}\]

The proof (Supplement S2) uses the fact that long sequences of omissions are increasingly explained by small values of the shared rate \(\mu\). We expand \(\log  m_{n}(\mu )\) at this boundary and rescale the integral by \(T=Cd_{n}\). The remainder can be controlled uniformly in \(n\), so the result applies when the number of contexts and the number of observations within each context grow together.

The law separates the contributions of the two forms of experience. At fixed \(n\), evidence increases as \(A \log  C+O(1)\) when the learner encounters additional contexts. At fixed \(C\), increasing \(n\) gives \(d_{n}=\kappa  \log  n+O(1)\), so the evidence increases only as \(A \log  \log  n+O(1)\). When both increase, the joint scale is approximately \(C\kappa  \log  n\) for large \(n\). For the special case of one observation in every context, the exact likelihood is \(L_{C,1}=B(A,B+C)/B(A,B)\).

The double-logarithmic growth at fixed context count reflects two possible low-rate explanations: the local context may be unfavorable to the expression, and the language environment as a whole may tend to use it rarely. Encountering additional contexts constrains the first explanation, while the second remains available through the shared mean. Thus the linear known-rate, logarithmic local-Beta, and shared-mean regimes differ in how much uncertainty the learner can resolve through repetition. Supplement S2 discusses further regimes, including positive lower bounds on use and point mass at zero. The asymptotic law describes long-run behavior; the following result asks how context allocation matters for any fixed finite number of omissions.

\subsection{2.5 Context refinement at fixed observation
count}\label{context-refinement-at-fixed-observation-count}

Imagine that the learner observes sixteen occasions without hearing the target. All sixteen may come from one persistent setting, or they may be divided among several settings with independently drawn local rates given the same shared mean. Because a low rate in a single setting can explain many omissions together, these two histories may carry different evidence even though their observation counts are identical. The following comparison makes this precise without requiring a Beta distribution for the local rate.

\textbf{Proposition 2 (context refinement).} Suppose a local rate is persistent and new contexts draw rates independently conditional on a common \(\mu\). For \(m_{n}(\mu )=\mathbb{E}[(1-\theta _{c})^n\mid \mu ]\) and positive integers \(a,b\),

\[m_{a+b}(\mu)\geq m_a(\mu)m_b(\mu).\tag{9}\label{eq:9}\]

To see why, let \(X=1-\theta _{c}\). Both \(X^a\) and \(X^b\) increase with \(X\), so their covariance is nonnegative. The inequality is strict when the conditional local-rate distribution is nondegenerate. Multiplying by the factors from other contexts and integrating over the same shared mean preserves the ordering (Supplement S3).

The left-hand side represents two observation blocks that share a single local rate; the right-hand side represents the same numbers of omissions in two independently drawn local contexts. A single low rate can account for both blocks on the left, whereas on the right the likelihood averages over separate local rates. Splitting a persistent block into conditionally independent contexts therefore lowers the probability of complete absence under permission, and increases its evidence against permission. We call this change in the generative allocation of observations \emph{context refinement}. The inequality applies to every conditional local-rate distribution on {[}0,1{]} and to successive refinements of the partition. It motivates experiments that hold exposure constant while varying the independence of the contexts in which it occurs.

\subsection{2.6 Positive observations calibrate later
absence}\label{positive-observations-calibrate-later-absence}

The preceding results concern learners who have observed omissions of a target expression. In real language experience, learners also hear other expressions and observe situations in which different meanings become relevant. If those experiences inform the expected use of the target, then two learners with identical target histories may interpret the target\textquotesingle s omissions differently. We examine two ways in which such calibration can occur: observing the use of related expressions, and observing semantic opportunities directly.

\textbf{Proposition 3 (shared-rate calibration).} Consider independent auxiliary indicators \(X_{j}\mid \mu \sim \operatorname{Bernoulli}(\mu )\) that sample the effective-use environment of the target hierarchy in Equation~\eqref{eq:6}. With \(k\) successes among \(M\) probes, the posterior for the shared mean is

\[\mu\mid X\sim\mathrm{Beta}(A+k,B+M-k),\qquad L(\mathbf n\mid k,M)=\mathbb E_{\mu\mid X}\!\left[\prod_c m_{n_c}(\mu)\right].\tag{10}\label{eq:10}\]

At fixed \(M\) and target observation allocation \(n\), this likelihood decreases as \(k\) increases. Frequent auxiliary successes shift beliefs toward higher shared use rates, making the same subsequent sequence of target omissions less probable under permission. Hence they increase the evidence against permission.

One way to obtain these auxiliary observations is to record other expressions that are known to be permitted and whose effective usage rates are exchangeable with the target\textquotesingle s rate conditional on \(\mu\). If each auxiliary observation samples a separate context, drawn independently of the other auxiliary and target contexts conditional on \(\mu\), the probability of encountering the auxiliary expression on that occasion is \(\mu\). Related expressions can then inform how often the target might be heard without requiring any occurrence of the target itself. In the factorized model where the initial permission probability is independent of the shared rate given \(S\), the auxiliary observations leave \(\pi (S)\) unchanged while calibrating the likelihood used in Equation~\eqref{eq:2}.

A second route to calibration begins with the communicative situation itself. By observing scenes in which the relevant meaning arises, learners can estimate how frequently opportunities for its expression become available. Let \(\nu\) be a shared mean for semantic opportunities, and suppose \(q_{c}\mid \nu \sim \operatorname{Beta}(\kappa \nu ,\kappa (1-\nu ))\). When the target is permitted, let its conditional form-choice probability be a known value \(r_{0}\). Auxiliary scene labels inform \(\nu\), and the probability of \(n\) target omissions within a context becomes \(\mathbb{E}[(1-r_{0}q_{c})^n\mid \nu ]\). More positive scene labels again strengthen the evidence in subsequent target omissions. When \(r_{0}=1\) this reduces to the effective-rate calculation in Equation~\eqref{eq:10}; when \(0<r_{0}<1\) it allows for the possibility that a relevant meaning arises but speakers choose an alternative (Supplement S4).

In the conjugate effective-rate model, the \(k\) positive auxiliary observations also change the low-rate behavior of the posterior. For fixed \(M\) and \(k\), the exponent A in Equation~\eqref{eq:8} becomes \(A+k\). Each positive auxiliary observation increases the coefficient of logarithmic evidence accumulation. Positive experience therefore influences the entire subsequent trajectory of restrictive learning, including its asymptotic rate of evidence accumulation. This interaction between expectations and later omissions has a further consequence for the relative plausibility of candidate expressions.

\subsection{2.7 Structural support can reverse its posterior
ordering}\label{structural-support-can-reverse-its-posterior-ordering}

Structural evidence can make one unfamiliar expression appear more plausible than another. Independent information about language use can also make the initially favored expression seem more likely to occur if permitted. In this situation, continued absence works against the stronger initial expectation. We ask when the resulting evidence can outweigh the initial structural advantage.

\textbf{Proposition 4 (support reversal).} Consider candidates \(H\) and \(L\) with \(0<\pi _{L}<\pi _{H}<1\) and known effective rates \(0\leq \theta _{L}<\theta _{H}<1\). After \(n\) omissions for each candidate, their posterior log-odds difference is

\[D_n=\operatorname{logit}\pi_H-\operatorname{logit}\pi_L+n\log\frac{1-\theta_H}{1-\theta_L}.\]

The difference starts positive and decreases linearly because each omission is less probable for \(H\) under permission than for \(L\). The posterior ordering strictly reverses when

\[n>n^*=\frac{\operatorname{logit}\pi_H-\operatorname{logit}\pi_L}{\log(1-\theta_L)-\log(1-\theta_H)}.\tag{11}\label{eq:11}\]

For example, let \(\pi _{H}=0.90\), \(\pi _{L}=0.60\), \(\theta _{H}=0.30\), and \(\theta _{L}=0.04\). Initially \(H\) is the more plausible candidate. Its omissions, however, provide considerably more evidence against permission than the corresponding omissions of \(L\). Here \(n^{*}=5.67\), and the ordering reverses after the sixth omission. The stronger expected use of \(H\) makes each omission more informative about its permission. The contrast in expected usage determines how rapidly the initial advantage diminishes, while the size of that advantage and the difference in omission evidence jointly determine the crossover point.

When the effective rates are uncertain, the learner averages each candidate's omission likelihood over its own distribution of possible usage rates. The number of contexts and the duration of experience within each context then help determine the cumulative evidence for each candidate (Supplement S5). The model thus predicts a learning trajectory in which an initial structural preference can gradually reverse as especially informative omissions accumulate. Observing such a trajectory requires measurements that distinguish the processes governing expected use.

\subsection{2.8 Observing semantic opportunities separates opportunity
from
choice}\label{observing-semantic-opportunities-separates-opportunity-from-choice}

Suppose a permitted expression occurs on 21\% of observation occasions. The relevant meaning might arise on 35\% of occasions, with speakers choosing the target 60\% of the time when it arises. Alternatively, the meaning might arise on 70\% of occasions, but speakers choose the target only 30\% of the time. Both environments produce a target occurrence probability of 0.21, even though the semantic opportunities and preferences for the expression differ considerably. These two environments are observationally equivalent when the data consist solely of target occurrences.

\textbf{Proposition 5 (observation and identification).} Conditional on permission, let \(q,r\in (0,1)\) be constant over \(N\) independent occasions. Target occurrences \(Y\) are Bernoulli with probability \(qr\). The Fisher information matrix describes how strongly observations constrain different directions in the parameter space. Its rank indicates how many independent parameter directions are locally identifiable. The Fisher information matrix for the parameters \(q\) and \(r\) is

\[\mathcal I_Y=\frac{N}{qr(1-qr)}\begin{bmatrix}r^2&qr\\qr&q^2\end{bmatrix}.\tag{12}\label{eq:12}\]

The matrix \(\mathcal I_Y\) has rank one, corresponding to a single identifiable combination of the two parameters. For any admissible positive \(c\), replacing \(q\) by \(cq\) and \(r\) by \(r/c\) preserves both the product \(qr\) and the distribution of target occurrences. Increasing the number of occurrence-only observations therefore improves the precision of the estimated product, while the individual parameters remain observationally equivalent along a curve of constant \(qr\).

If the researcher also records whether a relevant semantic opportunity arose, then an omitted target can be distinguished from an occasion on which its meaning was unavailable. Under the same assumptions, observing \(O\) and \(Y\) gives

\[\mathcal I_{O,Y}=\operatorname{diag}\left(\frac{N}{q(1-q)},\frac{Nq}{r(1-r)}\right),\tag{13}\label{eq:13}\]

The matrix \(\mathcal I_{O,Y}\) has full rank for interior values of \(q\) and \(r\). The opportunity rate \(q\) can be estimated from all observed scenes, and the conditional choice rate \(r\) from the scenes in which \(O=1\). Conditional on permission, scene-level annotations therefore make these two components of language use separately identifiable. In a broader learning experiment, structural cues, known permitted expressions, and subsequent production histories can also provide evidence about the permission state \(Z\) (Supplement S6). Together, these observations offer a way to study how learners acquire both structural knowledge and expectations about the use of permitted forms.

\section{3. Methods}\label{methods}

\subsection{3.1 Computational theory and artificial
grammars}\label{computational-theory-and-artificial-grammars}

We computed exact likelihoods and posterior probabilities using Beta-function ratios and finite products. We evaluated integrals over the shared mean using Gaussian quadrature and checked the results using adaptive quadrature. To examine how the theoretical results extend across different patterns of experience, we evaluated the joint evidence law for 75 combinations of prior parameters, context numbers, and within-context observation counts, and the context-refinement inequality for 140 combinations of parameters and exposure allocations. To examine calibration while holding the target history fixed, we presented the model with twelve auxiliary probes followed by sixteen target omissions, with four omissions in each of four contexts (4 \(\times\) 4). We evaluated both the effective-rate and semantic-opportunity hierarchies; the latter used a conditional choice probability \(r_{0}=0.60\). Support reversal was examined for known fixed rates (Section 2.7) and for uncertain hierarchical rates (Supplement S5). Finally, to determine how accurately observing opportunities permits estimation of \(q\) and \(r\), we simulated 4,000 datasets at each sample size \(N\in \{32,64,128,256,512\}\), with true \(q=0.35\) and \(r=0.60\).

We also examined whether the pattern of complete absence across exposure designs can distinguish the three usage-rate models. For this purpose, we simulated data under the known fixed-rate, persistent local-rate, and uncertain shared-mean models across eighteen observation-allocation designs. Supplement S12 describes the recovery of these generating mechanisms from the resulting presence-or-absence observations.

To examine whether shared continuations support generalization without direct pairwise experience, we generated thirty independently randomized artificial grammars. Each grammar contained three categories, twelve determiners, thirty-six nouns, and twelve continuation tokens. Three nouns in each category appeared only after a neutral initial frame, so these nouns had no direct training adjacency with any determiner. We evaluated whether SCB assigned higher compatibility scores to unseen determiner--noun pairs from the same category than to unseen pairs from different categories. Training sets contained 120, 300, 700, 1,500, or 4,000 sentences. To isolate the contribution of predictive structure, we permuted the continuation tokens after generation, preserving their marginal frequencies and determiner--noun exposure structure while disrupting the associations between those words and what followed them. We compared SCB with direct-pair memory and two alternative ways of extracting distributional structure: cosine similarity between continuation profiles and \(k\)-means clustering of the continuation profiles. We measured discrimination using AUC, which compares the ranks of compatible and incompatible pairs, assigning half credit for ties. Confidence intervals were obtained by resampling independent grammar seeds. Supplement S7 describes the grammar design and continuation-noise manipulation in detail, and examines the distinction between structural generalization and sequence recognition using gated recurrent unit (GRU) models and human sequence-recognition data.

\subsection{3.2 Meaning-conditioned judgments and linked
production}\label{meaning-conditioned-judgments-and-linked-production}

The artificial-language experiments of Samara et al. \cite{art02} provide a way to separate the effect of hearing a word frequently from the effect of hearing an alternative that conveys the same meaning. The experiments contrasted a preemption condition, in which two constructions expressed matching event meanings, with an entrenchment condition, in which they expressed distinct meanings. The utterances and exposure counts were matched between these conditions. Across the five experiments, the archive contains 440 participant records, comprising 12,760 judgment trials and 14,080 production trials. Our judgment analysis uses 10,792 semantically matched trials. The critical contrast involves two kinds of combinations that were not heard in training: combinations involving a novel word introduced at test, and combinations involving a familiar restricted word that had occurred only in the other construction. We also distinguish combinations attested in training and combinations involving alternating words that occurred in both constructions. For each participant, we calculated a restriction index as the mean rating of novel combinations minus the mean rating of unattested restricted combinations. A larger index indicates greater reluctance to extend a familiar restricted word relative to a novel word. The preemption-minus-entrenchment contrast was estimated within each experiment and then averaged with equal weight across experiments; participants were resampled within condition and experiment to obtain intervals.

We next asked whether judgments of linguistic restriction contain information about how the same learner chooses to speak. We linked each participant\textquotesingle s judgment profile to their production records. Critical production trials involved a restricted word, a scene calling for the construction absent from that word\textquotesingle s training history, and a response using one of the two marked constructions. The dependent variable was whether the learner produced the unattested restricted combination on such a trial. The resulting paired sample comprised 399 participants and 3,162 valid productions. Our reference prediction model included experiment-by-condition intercepts and ratings of novel and attested combinations, providing information about training condition and general judgment behavior. The extended model additionally included the rating of the unattested restricted combinations. To examine whether any extra judgment measure would improve prediction, we also compared a model that instead added ratings of alternating items. Models were fitted by binomial logistic regression to each participant\textquotesingle s restricted-choice count and total valid production count.

To examine whether the relationship between judgments and production extends to new learners, we used fivefold cross-validation, with participants grouped by experiment and training condition for fold assignment. We also withheld entire experiments in turn, using a common indicator of training condition to assess whether the relationship generalizes across experimental settings. Prediction was evaluated by the change in held-out log probability per production: a positive gain indicates that the additional judgment information improves the probability assigned to the learner's observed choices. Because learners contributed different numbers of productions, we report both the average gain across participants and the gain averaged over all eligible productions. The former gives every learner equal influence, and the latter gives every production equal influence. We estimated uncertainty using 20,000 bootstrap samples of participants. Supplement S8 provides the response definitions, sample composition, and scoring calculations.

Learners in different training conditions may also differ in how they use the five-point acceptability scale. We examined these differences with an ordinal model that allows each condition to have its own thresholds separating successive rating categories. The model also includes a condition effect specific to unattested restricted combinations, enabling us to examine how much additional information is carried by changes in other item classes. We evaluated these models using eight participant folds within each experiment.

The original experiments involved different numbers of exposures to restricted words: 64 in Experiments 1--2, 128 in Experiment 3, and 56 in Experiments 4--5. Exposure count was fixed within each experiment, making it inseparable from differences between experimental designs. These experiments provide a controlled setting for studying meaning-matched competition and the relationship between judgments and production. To isolate the effect of context allocation predicted by Proposition 2, an experimental design would distribute a fixed number of exposures across different communicative contexts within the same experiment.

\subsection{3.3 Natural language use across families and
corpora}\label{natural-language-use-across-families-and-corpora}

The hierarchical theory assumes that the rate of encountering a linguistic form may differ across persistent communicative environments. To examine this assumption in natural speech, we used English article use as a measurable example. The noun \emph{cake}, for instance, can occur with an article in \emph{make a cake} or without one in \emph{have cake}. Our question is whether such differences are systematically associated with the source of the speech even when the lexical item and major grammatical properties are held comparable. We combined the gold-annotated utterances from the training, development, and test partitions of UD-English-CHILDES \cite{art08}. Adult speech includes mothers, fathers, investigators, grandparents, caretakers, and relatives; child speech is that of the target child. We coded each common-noun occurrence according to whether it was accompanied by an article (\(Y_t=1\)). An article was identified as a dependent of the noun with the dependency relation \texttt{det}, the part-of-speech tag \texttt{DET}, and the lemma \emph{a} or \emph{the}; the lemma \emph{a} also includes \emph{an}. To compare article use under similar grammatical conditions, we grouped noun occurrences by lexical identity, grammatical number (singular \texttt{NN}, plural \texttt{NNS}, or unspecified), and syntactic role (subject \texttt{nsubj}/\texttt{csubj}, object \texttt{obj}/\texttt{iobj}, syntactic root \texttt{root}, or other). Each resulting group therefore contains occurrences of the same noun under comparable grammatical conditions.

We compared two models that share the same lexical and grammatical information but differ in their treatment of recording sources. The reference model integrates uncertainty about lexical article rates and about rates conditional on grammatical number and dependency role. The source-varying model further allows each source to have its own rate drawn around the uncertain reference mean. We estimated the shared priors and concentration parameters from adult training speech. Comparisons among training sources determined how strongly individual source rates were drawn toward the common reference distribution. Nouns were eligible for evaluation if they occurred at least eight times in the adult training data. To assess prediction across new recording sources, we fitted the models after withholding the evaluation source in its entirety. We examined the original division into seven training families and four test families, all eleven leave-one-family-out evaluations, and all six leave-one-corpus-out evaluations. In the last design, every family belonging to the evaluation corpus was withheld together.

We evaluated each model by the joint marginal log probability assigned to the observations from a held-out source, expressed per noun token. To estimate uncertainty, we resampled families or corpora together with their total log-score gains and numbers of observations. This preserves the contribution of each evaluated token in the weighted estimates and their uncertainty intervals. We also report averages that give each recording source equal weight. A separate conditional randomization analysis examined whether the observed concentration of absences across sources could arise from a common use rate alone. Within each word-by-number-by-role stratum, we held fixed the total article count and the number of noun tokens contributed by each family, and randomly reallocated the articles 1,999 times. We compared the observed number of source cells containing no articles and a measure of variation in article use across families within each stratum (within-stratum dispersion) with their corresponding randomization distributions. This provides a direct description of usage clustering alongside the held-out model comparison. Supplement S9 presents the original split and the cross-source analyses.

\subsection{3.4 Transfer of use rates and five-language model
comparison}\label{transfer-of-use-rates-and-five-language-model-comparison}

Shared predictive continuations can provide evidence about structural compatibility. Estimating how frequently a form is used in natural language raises a further question about the information available in these continuations. We examined whether the words immediately following a noun, considering the first and second following positions, help predict article-use rates for nouns whose article-use observations were withheld during training. We evaluated predictions across five sets of held-out nouns, using four inner cross-validation folds to select model settings. Supplement S10.4 examines the predictive contribution of these continuation features through permutations among nouns of similar frequency. As a complementary representation, we used dependency profiles, which describe the distribution of a noun's syntactic dependency relations. Each profile gives the proportions of subject, object, root, and other occurrences in adult training speech. The article labels of the evaluation nouns were reserved for the final prediction assessment. Supplement S10 describes this analysis together with the analyses of complete-absence probabilities and temporal ordering.

The choice among causative constructions provides another way to study the information needed to predict form preferences. Speakers can often describe a causal event in different ways; for example, English allows \emph{She broke the vase} and \emph{She made the vase break}. Languages vary in how causative meaning is expressed and in the lexical and semantic properties associated with each form. The archive of Ambridge et al. \cite{art09} includes 13,169 usable child judgments, for sixty verbs in each of English, Hebrew, Hindi, Japanese, and K'iche', as well as lexical construction counts and four semantic ratings. We used five folds within each language, holding out all behavioral judgments for the verbs in each test fold. The corpus counts and semantic features of those verbs remained available to every model: the task was to predict previously unseen behavioral items from an available lexical-input corpus. Frequency-smoothing priors, behavioral coefficients, and feature scaling were estimated using training verbs only.

We compared seven models that use different combinations of construction form, lexical frequency, and semantic information. These included models based on form, frequency, or semantics individually, an additive model combining frequency and semantics, and models based on Beta-smoothed lexical rates with or without semantic features. We also implemented the softmax architecture developed in the original study \cite{art09}. This model learns a distribution over competing linguistic forms from lexical identity, causative marking, and semantic information. Its sixty lexical inputs, one causative input, and four semantic inputs connect directly to three mutually exclusive outputs, together with bias weights.

Following the original modeling protocol, we trained the network on forty-eight independently sampled corpora of 10,000 tokens by minimizing a regularized cross-entropy objective. To connect the corpus predictions with children's judgments, we used a logistic model whose predictors were the logit-transformed activation of the judged form and an indicator of that form. This behavioral model was fitted on the same training verbs used for the other models. All seven models were consequently evaluated on a common task: predicting children's judgments of held-out verbs. Supplement S11 presents the mathematical objective, optimization procedure, and remaining training details.

To characterize prediction across lexical items, we first averaged the log-score gains within each verb and then averaged across verbs. This gives each verb equal influence in the evaluation. Uncertainty intervals were estimated using 20,000 bootstrap samples of verbs. Supplement S11 presents the full model comparisons and the corpus information available to each model.

\begin{figure}[!htbp]
\makebox[\linewidth][c]{%
\begin{minipage}{166mm}
\captionsetup{font=integrated,skip=6pt}
\centering
\includegraphics[width=166mm,height=228mm,keepaspectratio]{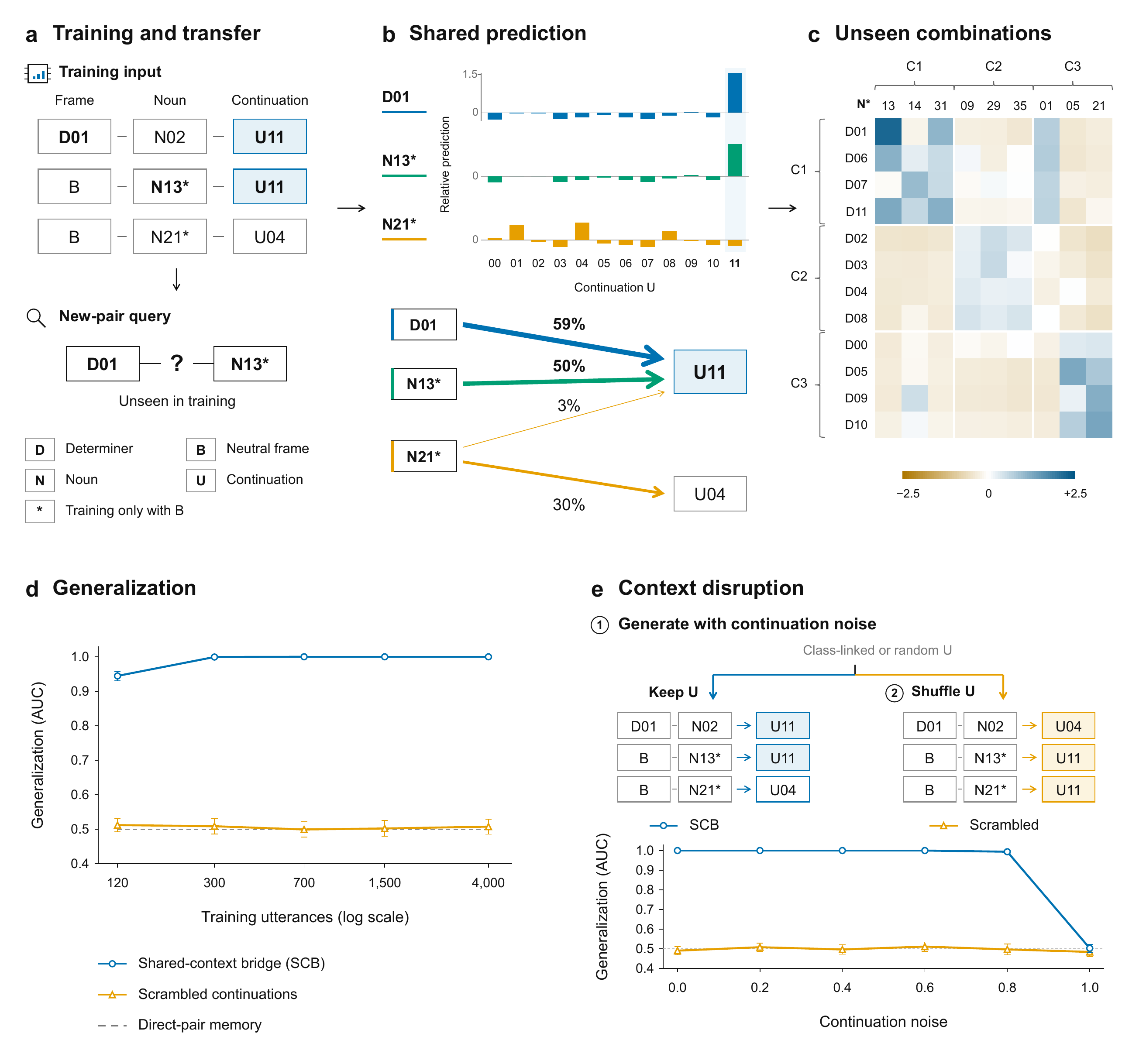}
\caption{\textbf{Shared predictions provide evidence for previously unseen combinations.} \textbf{(a)} The training utterances contain determiners (D), nouns (N), a neutral initial frame (B), and continuation tokens (U). Nouns marked N* appear only in neutral frames; the D--N* combination queried at test has never occurred in training. \textbf{(b)} The determiner D01 and noun N13* both predict continuation U11. The upper panel shows their centered, frequency-weighted continuation profiles; the arrows below show selected conditional probabilities, with width proportional to probability. \textbf{(c)} SCB scores for all 108 unseen determiner--noun pairs, ordered by the generator\textquotesingle s three latent categories. \textbf{(d)} Discrimination of compatible and incompatible pairs as exposure increases, comparing SCB, the same estimator after scrambling continuations, and direct-pair memory. Points and intervals are means and 95\% grammar-bootstrap intervals over 30 generated grammars. AUC is a rank-based measure of discrimination, with chance performance at 0.5. \textbf{(e)} Category-independent continuation noise weakens the predictive signal during generation. In the scrambled condition, continuations are additionally permuted across utterances after generation, preserving their marginal frequencies and word-pair exposure structure. Curves show the two SCB conditions at 4,000 training utterances. The noise fraction in panels a--d is 0.2.}
\label{fig:1}
\end{minipage}%
}
\end{figure}

\section{4. Results}\label{results}

\subsection{4.1 Shared continuations support new structural
combinations}\label{shared-continuations-support-new-structural-combinations}

We first asked whether two words could be recognized as structurally compatible even if they had never appeared together. Figure~\ref{fig:1}a--b illustrates how this can be achieved: the determiner D01 and the noun N13* are never paired during training, yet both frequently predict continuation U11. Their separate histories therefore supply a potential basis for generalization. Across thirty randomly generated grammars, SCB assigned higher compatibility scores to unseen pairs from the same category than to unseen pairs from different categories after only 120 sentences (mean AUC 0.945, 95\% CI \([0.931,0.957]\); Figure~\ref{fig:1}d). Because the training adjacency count was zero for all evaluated pairs, direct-pair memory could not distinguish the categories (AUC 0.500). At the continuation-noise level used in Figure~\ref{fig:1}d (\(\epsilon=0.20\)), SCB reached an AUC of 1.000 after 700 training sentences.

To determine whether the common predictive information was responsible for this performance, we disrupted it while preserving lexical marginals and determiner--noun exposure structure. Permuting the continuations reduced AUC at 120 sentences to 0.512 \([0.493,0.531]\). Varying the category-independent continuation noise also changed the strength of the structural signal in the manner predicted by Equation~\eqref{eq:4}. For \(\epsilon =0.20\), the analytical within-minus-between-category margin was 1.920 and the numerical estimate 1.903; for \(\epsilon =0.80\), the corresponding values were 0.120 and 0.111 (Figure~\ref{fig:1}e and Supplement S7). Two alternative ways of extracting distributional structure, cosine similarity between continuation profiles and \(k\)-means clustering of continuation profiles into categories, also distinguished compatible from incompatible combinations. Together, these results reveal how a learner could acquire structural evidence for a novel combination from the separate histories of its component words. Once such evidence has established an initial expectation of compatibility, subsequent experience can refine that expectation. We next examine what happens when the supported combination remains absent.

\begin{figure}[!htbp]
\makebox[\linewidth][c]{%
\begin{minipage}{166mm}
\captionsetup{font=integrated,skip=6pt}
\centering
\includegraphics[width=166mm,height=228mm,keepaspectratio]{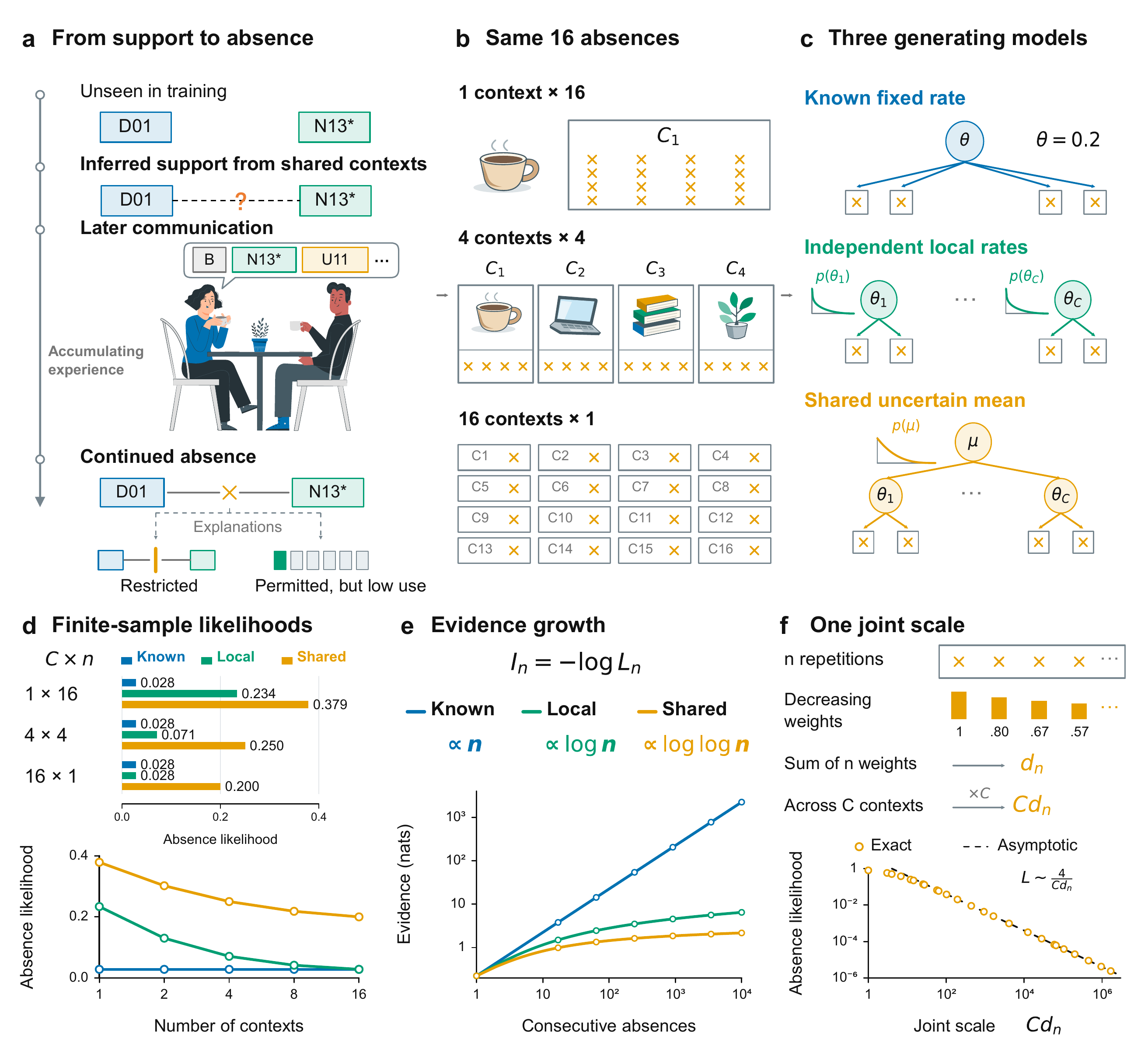}
\caption{\textbf{The evidence in an absence depends on how observations are distributed across contexts.} \textbf{(a)} Shared predictions first support a candidate combination that was absent during training. Subsequent experience in which the target remains absent can be explained by restriction or by low use despite permission. \textbf{(b)} The same sixteen target omissions are allocated to one, four, or sixteen contexts, with a persistent local usage rate within each context. \textbf{(c)} Three generating models: a known fixed rate \(\theta =0.20\); independent local rates \(\theta _{c}\sim \operatorname{Beta}(0.8,3.2)\); and an uncertain shared mean \(\mu \sim \operatorname{Beta}(1,4)\), with \(\theta _{c}\mid \mu \sim \operatorname{Beta}(4\mu ,4(1-\mu ))\). \textbf{(d)} Exact likelihoods of complete absence under permission. A lower likelihood means stronger evidence against permission; bar and curve colors denote the same models. \textbf{(e)} Evidence \(I=-\log  L\) in a single persistent context grows asymptotically at linear, logarithmic, or double-logarithmic rates under the three respective assumptions. \textbf{(f)} Within-context observations contribute decreasing weights, whose sum is \(d_{n}\). Across C conditionally independent contexts the joint scale is \(Cd_{n}\). In the displayed shared-mean hierarchy, the exact likelihood approaches \(4/(Cd_{n})\). Conversation artwork: Storyset/Freepik.}
\label{fig:2}
\end{minipage}%
}
\end{figure}

\subsection{4.2 Context diversity changes the evidence carried by the same number of omissions}\label{opportunity-organization-changes-the-evidence-in-equal-observation-counts}

Suppose a learner encounters sixteen observation occasions without hearing a target expression. All sixteen observations might arise in one persistent context, or each might arise in a different context. Should these histories carry the same evidence against permission? Figure~\ref{fig:2}a--c compares three accounts of how the observations are generated. In the simplest case, where the effective rate is known to be \(\theta =0.20\), all allocations give the same likelihood of sixteen omissions (\(L=0.028\)). A known usage rate gives every observation the same expected probability of target occurrence, so each omission contributes the same amount of evidence in every context.

The result changes when each context has its own uncertain usage rate. With independent local rates \(\theta _{c}\sim \operatorname{Beta}(0.8,3.2)\), sixteen omissions in one context have likelihood \(L=0.234\). A learner can explain the entire sequence through a low rate in that one context. When the same omissions are spread across sixteen independently sampled contexts, the likelihood falls to 0.028 (Figure~\ref{fig:2}d): multiple local rates must now account for the observed absence. An intermediate allocation of four contexts with four observations each gives \(L=0.071\). Finally, suppose the context-specific rates share an uncertain global mean \(\mu \sim \operatorname{Beta}(1,4)\), with \(\kappa =4\). The likelihood falls from 0.379 for one context to 0.250 for four contexts and 0.200 for sixteen. Context diversity continues to strengthen the evidence, while the possibility of a low shared mean provides an explanation that can extend across settings.

These comparisons illustrate the finite-sample ordering in Proposition 2 and the joint accumulation scale \(Cd_n\) in Proposition 1 (Figure~\ref{fig:2}e--f). Within a persistent context, each omission strengthens the explanation that its local usage rate is low, making subsequent omissions progressively more expected. This produces the decreasing weights in \(d_n\), which grows as \(\kappa\log n\) for large \(n\). Encountering a context with an independently drawn local rate supplies a new source of information, increasing the joint scale through \(C\). The numerical likelihoods approach the asymptotic expression across the conditions examined in Supplement S2. The evidential value of context diversity thus arises from the way successive observations sample the underlying distribution of usage rates.

\begin{figure}[!htbp]
\makebox[\linewidth][c]{%
\begin{minipage}{166mm}
\captionsetup{font=integrated,skip=6pt}
\centering
\includegraphics[width=166mm,height=228mm,keepaspectratio]{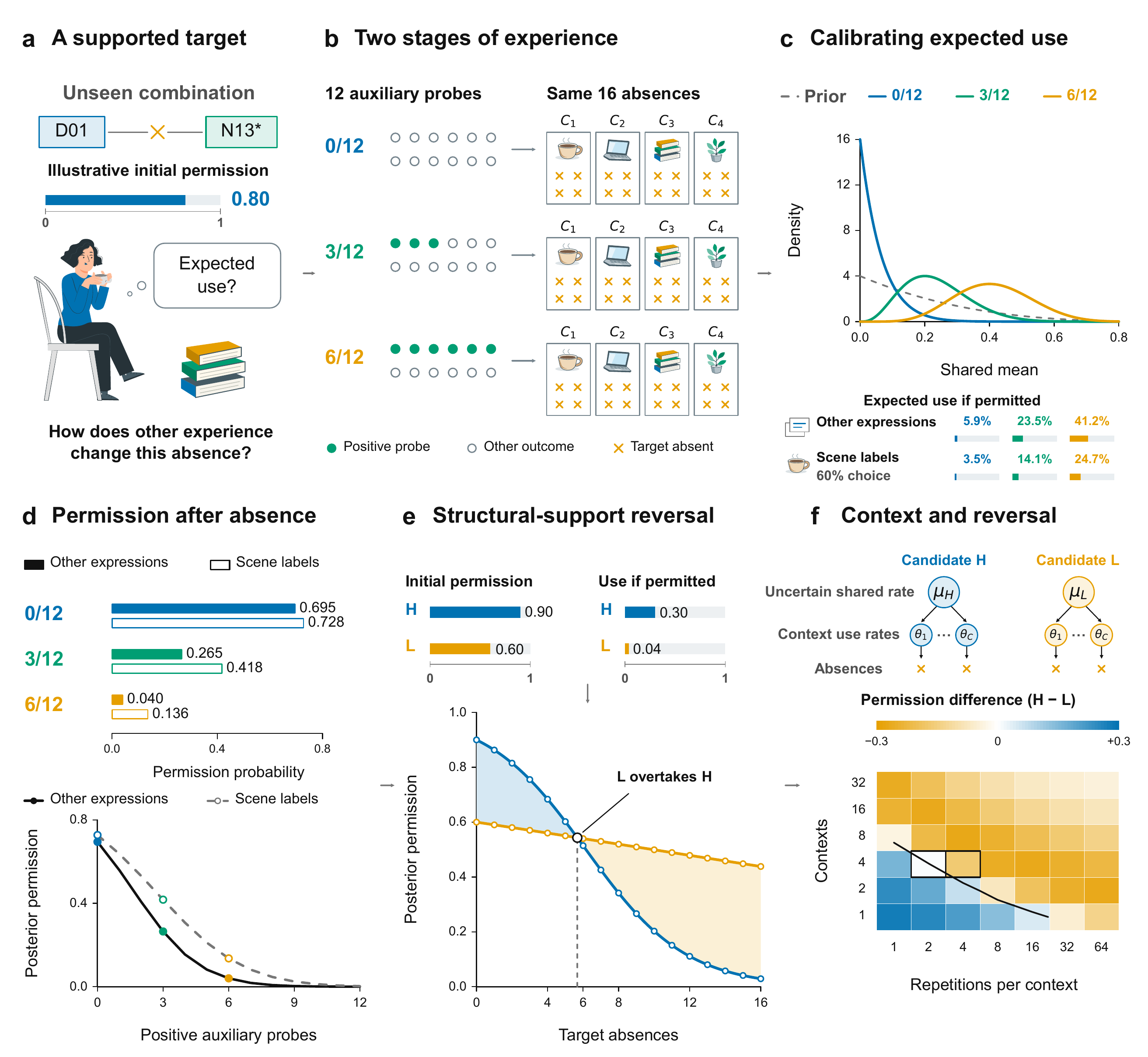}
\caption{\textbf{Positive experience changes the evidence in later omissions and can lead to a reversal of initial structural support.} \textbf{(a)} An unseen candidate receives initial structural support, illustrated by a permission prior of 0.80. \textbf{(b)} The model receives twelve auxiliary probes with zero, three, or six positive outcomes, followed by the same sixteen target omissions (four in each of four contexts). \textbf{(c)} The probes update the shared-rate distribution and the expected use of the permitted target. Related-expression probes inform the effective usage rate; scene labels inform the rate of semantic opportunities, with conditional choice probability 0.60. \textbf{(d)} Posterior permission after the fixed omission history as a function of the number of positive probes. Filled/solid and outlined/dashed marks distinguish the effective-rate and scene-label observation models. \textbf{(e)} Candidate H initially has higher permission than L (0.90 versus 0.60) but also higher expected use if permitted (0.30 versus 0.04). Omissions reduce H\textquotesingle s posterior more rapidly, reversing the ordering after six omissions. \textbf{(f)} Under shared-mean rate uncertainty, the posterior permission difference depends on the number of contexts and observations within each. Blue and amber denote positive and negative differences \(H-L\), respectively; the contour marks equality. Learner artwork: Storyset/Freepik.}
\label{fig:3}
\end{minipage}%
}
\end{figure}

\subsection{4.3 Positive experience calibrates subsequent
omissions}\label{positive-experience-calibrates-subsequent-omissions}

The preceding comparison changes where omissions occur. We next hold the target history fixed and change what the learner has previously discovered about the environment. Consider learners with the same initial belief that a new combination is permitted (\(\pi =0.80\)). Each receives twelve auxiliary probes and then observes sixteen target omissions, four in each of four contexts (Figure~\ref{fig:3}a--b). The probes differ in how often they provide positive evidence of effective use. Under the hierarchy with \(A=1\), \(B=4\), and \(\kappa =4\), no successes among the twelve probes suggest that relevant expressions may be rarely used. Subsequent target omissions reduce the permission probability only to 0.695. By contrast, three positive probes shift the shared-rate distribution toward more frequent use, and the identical target omissions reduce permission to 0.265. Six positive probes yield a posterior of 0.040 (Figure~\ref{fig:3}c--d). The stronger evidence arises from the changed expectation of use, even though the target\textquotesingle s structural prior and omission history are held constant.

A learner can also acquire information about usage opportunities without hearing a related expression, by observing whether the target meaning is relevant to a scene. We repeated the calculation with scene labels that update the shared semantic-opportunity mean and a conditional choice probability \(r_{0}=0.60\). Here zero, three, and six positive labels yield posterior permissions of 0.728, 0.418, and 0.136, respectively (Figure~\ref{fig:3}d). The conditional choice probability determines how strongly knowledge about relevant scenes influences the expected occurrence of the target. Because speakers may select alternative expressions, scene-based calibration produces a more gradual update than direct calibration of the effective usage rate in this example. Both routes reveal the same underlying principle: positive experience that increases expected target use also increases the evidence carried by subsequent omissions.

\subsection{4.4 Initially stronger support can produce stronger later
restriction}\label{initially-stronger-support-can-produce-stronger-later-restriction}

Calibration can also change the relative ordering of candidate expressions. Consider the two candidates illustrated in Figure~\ref{fig:3}e. Before encountering the omission sequence, the learner is more confident that \(H\) is permitted than that \(L\) is permitted (\(\pi _{H}=0.90\) versus \(\pi _{L}=0.60\)). But \(H\) also has a higher expected occurrence probability conditional on permission (\(\theta _{H}=0.30\) versus \(\theta _{L}=0.04\)). Consequently, observing an omission of \(H\) is more surprising than observing an omission of \(L\). The evidence difference eventually outweighs \(H\)\textquotesingle s initial advantage: the fixed-rate crossover occurs at \(n^{*}=5.67\), so \(L\) has the higher posterior permission probability after the sixth omission (Proposition 4).

In the hierarchical model, the rates are uncertain and each candidate has its own shared-rate distribution (Supplement S5). The number of contexts and the amount of repetition within them now influence the relative posterior probabilities through the integrated omission likelihoods. Figure~\ref{fig:3}f shows the resulting surface. For example, with four contexts and two omissions per context (eight in total), \(H\) retains a small posterior advantage of about 0.006. With four omissions per context (sixteen in total), the difference is \(-0.166\), favoring \(L\). The resulting learning trajectory reflects two interacting influences: structural evidence establishes an initial preference, while expected use determines how strongly subsequent omissions alter that preference. When the favored candidate is expected to occur sufficiently often, the cumulative evidence can reverse the initial ordering. Under hierarchical uncertainty, the timing of this reversal also depends on how the observations are distributed across contexts.

\begin{figure}[!htbp]
\makebox[\linewidth][c]{%
\begin{minipage}{159mm}
\captionsetup{font=integrated,skip=6pt}
\centering
\includegraphics[width=159mm,height=228mm,keepaspectratio]{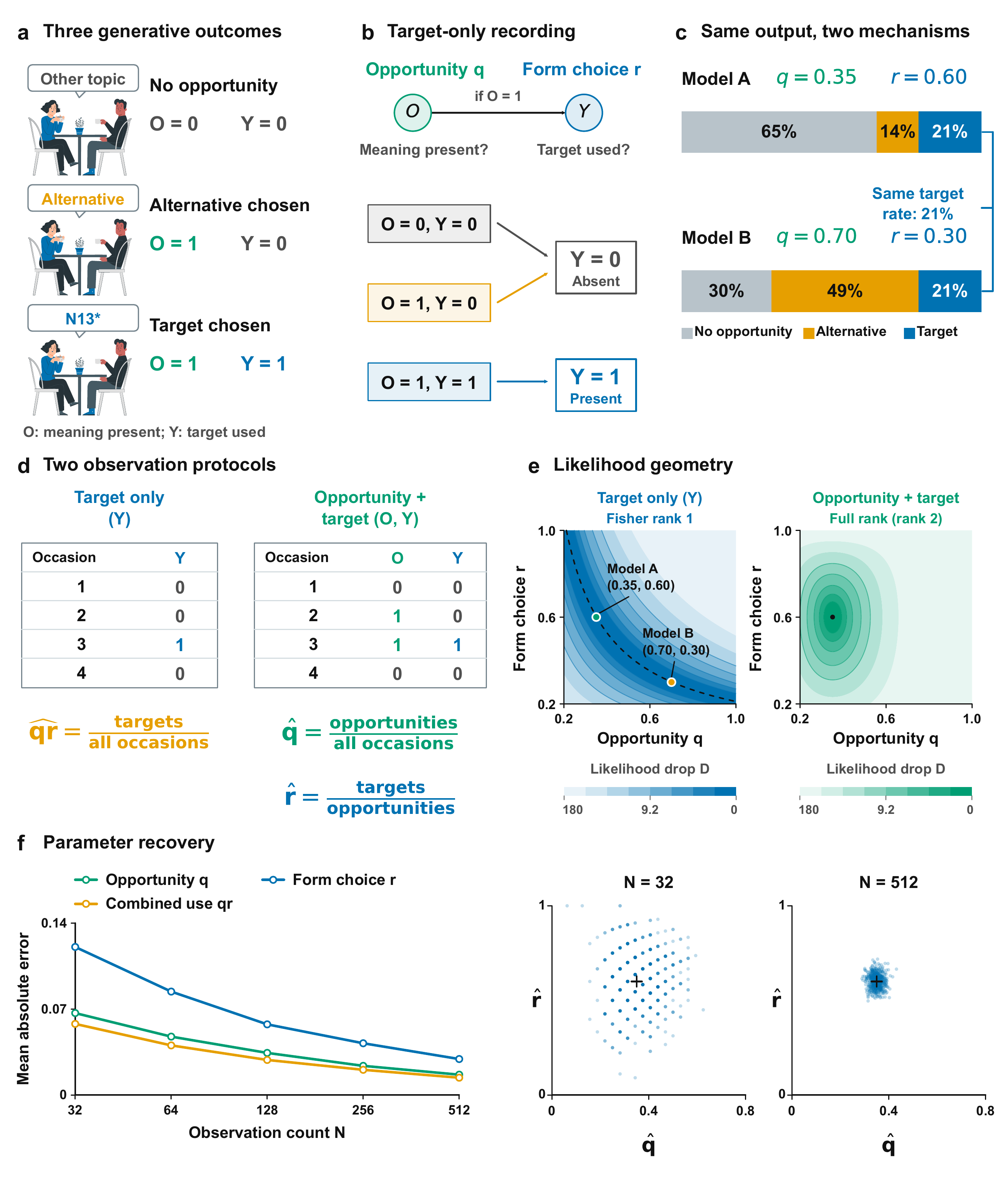}
\caption{\textbf{Recording semantic opportunities makes it possible to distinguish them from form choice.} All panels condition on the target being permitted. \textbf{(a)} The target can be absent because its meaning is irrelevant to the scene (\(O=0,Y=0\)) or because a speaker uses an alternative in a relevant scene (\(O=1,Y=0\)). It occurs when a relevant opportunity leads to selection of the target (\(O=1,Y=1\)). \textbf{(b)} Recording only target occurrences combines the two kinds of absence into \(Y=0\) and estimates the product qr. \textbf{(c)} Models A (\(q=0.35,r=0.60\)) and B (\(q=0.70,r=0.30\)) illustrate distinct opportunity and choice rates with identical target occurrence probability 0.21. \textbf{(d)} An observation record containing both O and Y distinguishes the counts needed to estimate q and r. \textbf{(e)} Likelihood geometry for \(N=128\) observations, including 45 opportunities and 27 target selections. Target-only observations give a rank-one Fisher information matrix and a likelihood ridge along parameter pairs with the same product \(qr\) (dashed line). Adding opportunity labels gives a full-rank information matrix and an isolated likelihood maximum for the illustrated observations. Shading shows twice the log-likelihood loss, with darker colors indicating higher likelihood. Points A and B correspond to panel c. \textbf{(f)} Parameter recovery in 4,000 datasets per sample size: mean absolute errors for q, r, and qr, and sampled estimates at \(N=32\) and \(N=512\). Crosses mark the true parameters (0.35,0.60). Conversation artwork: Storyset/Freepik.}
\label{fig:4}
\end{minipage}%
}
\end{figure}

\subsection{4.5 Opportunity observations separate two learning
processes}\label{opportunity-observations-separate-two-learning-processes}

The preceding results depend on the expected frequency of a target expression, which combines the availability of its meaning with the probability of choosing that form. Figure~\ref{fig:4}a--c illustrates two ways in which the same target occurrence rate can arise. A permitted expression may be selected on 60\% of occasions when its meaning is relevant, with such opportunities arising on 35\% of all occasions. Alternatively, opportunities may arise on 70\% of occasions, while speakers select the target on 30\% of those occasions. Both environments produce the same overall target occurrence probability of 0.21.

This equivalence appears directly in the geometry of the likelihood. At \(q=0.35\), \(r=0.60\), and \(N=128\), the occurrence-only Fisher information matrix has eigenvalues 0 and 372.27. The positive eigenvalue reflects the information available about the product \(qr\), while the zero eigenvalue corresponds to a direction along which changes in opportunity and choice compensate for one another. The likelihood consequently forms a ridge of equivalent parameter pairs (Figure~\ref{fig:4}e, left).

Recording whether each scene contains a relevant opportunity changes the information available to the researcher. Opportunities can now be counted separately from selections made within those opportunities (Figure~\ref{fig:4}d). At the same parameter values, the Fisher information matrix has two positive eigenvalues, 186.67 and 562.64. In the illustrated dataset, the likelihood has an isolated maximum that identifies a particular pair of opportunity and choice rates (Figure~\ref{fig:4}e, right).

The estimates become increasingly precise as more labeled scenes are observed. Across 4,000 simulated datasets at each sample size, the mean absolute errors for \(q\) and \(r\) decrease from 0.067 and 0.121 at \(N=32\) to 0.017 and 0.029 at \(N=512\) (Figure~\ref{fig:4}f). Conditional on permission, these results demonstrate how scene-level observations make separate estimation of semantic availability and conditional form choice possible.

\begin{figure}[!htbp]
\makebox[\linewidth][c]{%
\begin{minipage}{165mm}
\captionsetup{font=integrated,skip=6pt}
\centering
\includegraphics[width=165mm,height=228mm,keepaspectratio]{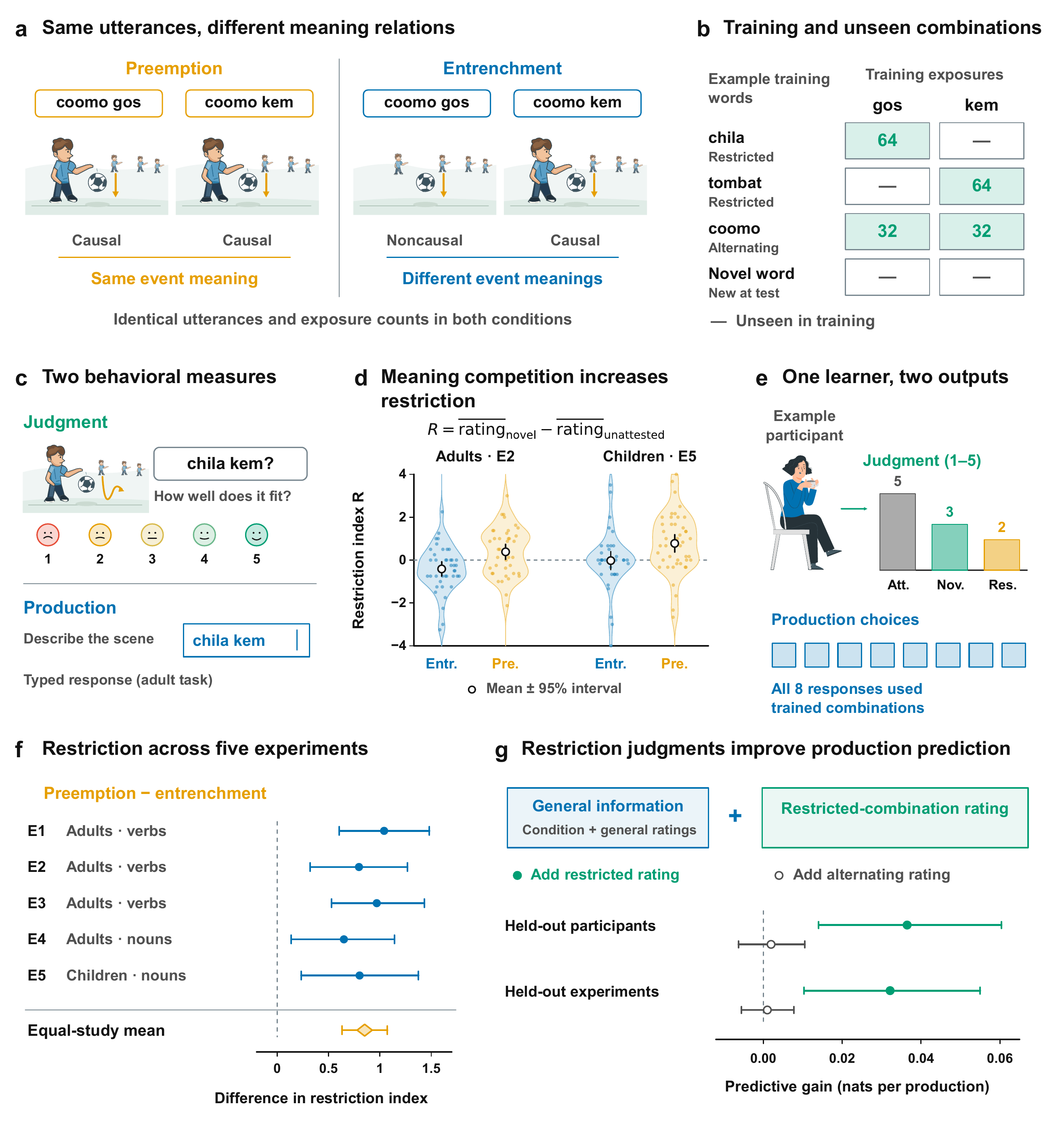}
\caption{\textbf{Meaning-matched competition increases restriction, and restricted judgments predict production choices.} \textbf{(a)} In Samara et al.\textquotesingle s randomized artificial-language experiments \cite{art02}, utterances and exposure counts are matched between conditions. The competing expressions convey the same event meaning under preemption and distinct meanings under entrenchment. \textbf{(b)} Example training words and their exposure counts; a dash indicates that the combination was not presented. The novel word is introduced only at test. \textbf{(c)} Learners judge the acceptability of expressions and describe scenes in separate behavioral tasks; the production example depicts the adult typed-response task. \textbf{(d)} Participant restriction indices (novel minus unattested restricted ratings) for adult Experiment 2 and child Experiment 5, with descriptive distributions and means/95\% intervals. \textbf{(e)} Example of one archived participant\textquotesingle s linked judgment and production profile: all eight critical productions use trained combinations. \textbf{(f)} Preemption-minus-entrenchment differences in restriction across the five experiments, with participant-bootstrap intervals and the experiment-equal mean (diamond). \textbf{(g)} Additional held-out production log-score gain when the reference model receives restricted-combination ratings or, as a specificity comparison, alternating-item ratings. Both participant and experiment holdouts are shown; points and bootstrap intervals weight participants equally (399 learners, 3,162 productions). Panels d and f summarize the original randomized training-condition comparisons, and panel g presents our analysis linking learners' judgments to their production choices. Learner artwork: Storyset/Freepik.}
\label{fig:5}
\end{minipage}%
}
\end{figure}

\subsection{4.6 Meaning competition and restriction information across
behavioral
tasks}\label{meaning-competition-and-restriction-information-across-behavioral-tasks}

We next examined whether human learners are especially reluctant to generalize when they repeatedly hear an expression that serves the same communicative function as a missing form. Samara et al.\textquotesingle s five artificial-language experiments \cite{art02} compare two groups that heard the same utterances with the same frequencies but experienced different relationships between their meanings (Figure~\ref{fig:5}a--c). In the preemption condition, the constructions competed to express matching meanings; in the entrenchment condition, they expressed distinct meanings. Across the five experiments, the restriction index (rating of novel combinations minus rating of unattested restricted combinations) was on average 0.852 rating points higher in the preemption condition (95\% CI \([0.632,1.073]\)). The differences in the individual experiments were 1.041, 0.797, 0.970, 0.650, and 0.803 (Figure~\ref{fig:5}d,f). Because the two conditions involved the same expressions and exposure counts, the stronger restriction under preemption reveals the contribution of competition between forms that express the same meaning.

Linguistic restriction can be expressed both in judgments about an unfamiliar combination and in the learner\textquotesingle s own choice of what to say. If these tasks draw on shared construction-specific knowledge, judgments of restricted items should help predict whether the same learner produces those items. We examined this by comparing a model of production using training condition and general ratings with a model additionally given each learner\textquotesingle s rating of the unattested restricted combinations (Figure~\ref{fig:5}e,g). For 399 participants and 3,162 valid productions, the additional information improved predictions on held-out learners by 0.0365 nats per production (participant-equal mean, 95\% CI \([0.0140,0.0604]\); production-weighted gain 0.0359 \([0.0135,0.0598]\)). Knowledge expressed in a learner's judgments therefore helps predict the linguistic choices that the same learner makes during production. The relationship also generalized across entire held-out experiments (+0.0322 nats \([0.0103,0.0550]\); production-weighted +0.0316 \([0.0099,0.0543]\)), with positive experiment-held-out gains in all five experiments (Supplement S8).

We further examined which aspects of a learner's judgments contribute to production prediction. Ratings of alternating items provide information about the learner's general response tendencies and therefore offer a useful comparison with ratings of restricted combinations. Adding alternating-item ratings yields a gain of 0.0020 nats per production, with a 95\% interval of \([-0.0063,0.0105]\) (Figure~\ref{fig:5}g). The larger gain from restricted-combination ratings indicates that the relevant information is closely associated with the particular restriction being learned.

A related question concerns how learners use the rating scale itself. In the ordinal analysis, each training condition was allowed its own thresholds separating the five rating categories, alongside an effect specific to unattested restricted combinations. Allowing additional condition effects for novel and alternating items improved prediction by only 0.0020 nats per judgment, with a 95\% interval of \([-0.0008,0.0048]\). Together, these comparisons reveal a shared, item-class-specific component of linguistic restriction across acceptability judgments and production choices. The generative framework represents the underlying knowledge through structural permission \(Z\) and conditional form choice \(r\), providing a basis for joint behavioral models in which both quantities contribute to the learner's judgments and choices.

\begin{figure}[!htbp]
\makebox[\linewidth][c]{%
\begin{minipage}{163mm}
\captionsetup{font=integrated,skip=6pt}
\centering
\includegraphics[width=163mm,height=228mm,keepaspectratio]{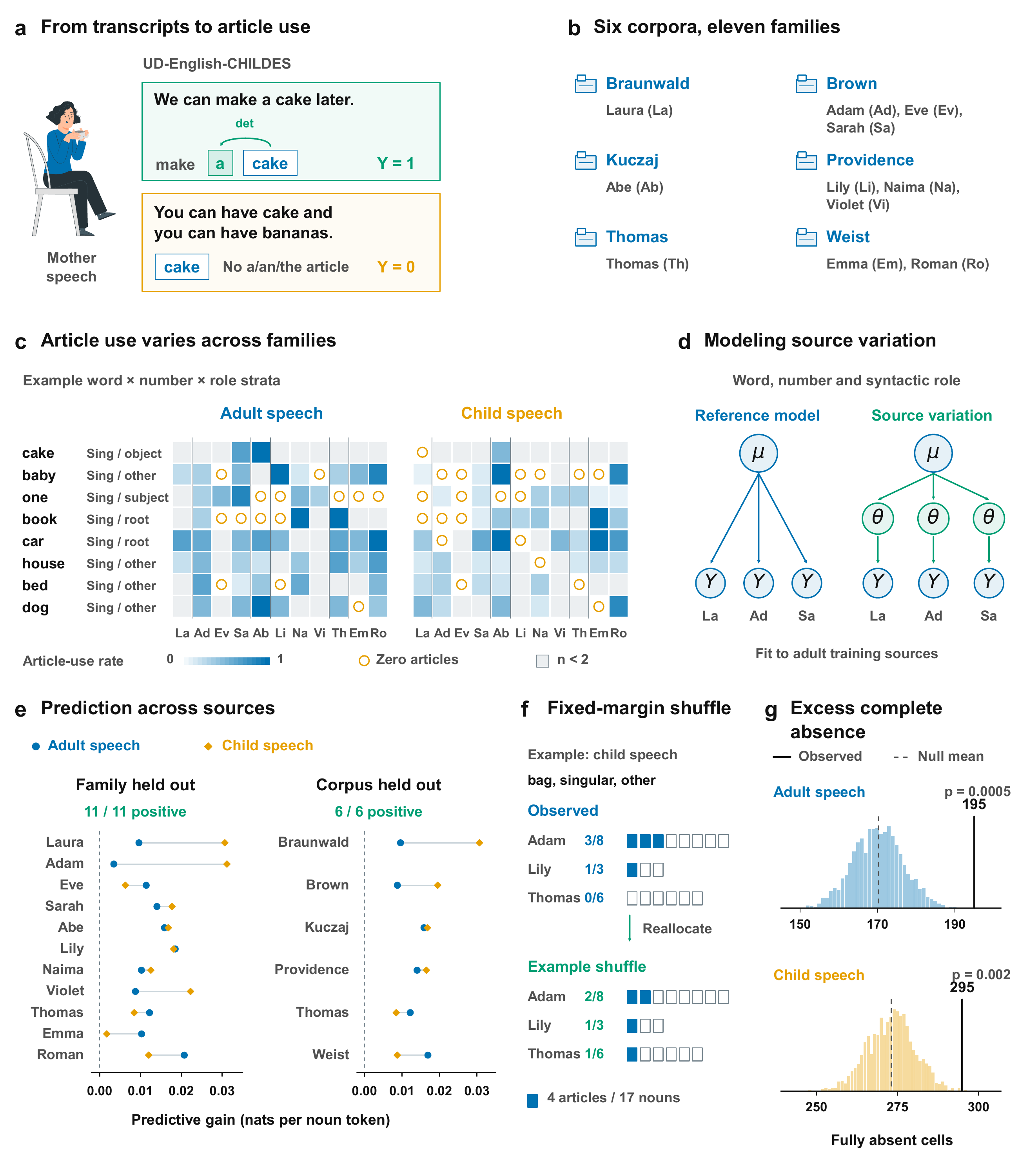}
\caption{\textbf{Article-use rates vary systematically across natural-language sources.} \textbf{(a)} Examples from annotated CHILDES transcripts show how a noun token is coded as occurring with or without an article (\emph{a/an/the}). \textbf{(b)} The eleven families represented in the six recording corpora. \textbf{(c)} Illustrative word-by-number-by-dependency-role strata for adult and child speech. Color denotes the article-use proportion; orange rings indicate cells with zero articles, and grey denotes fewer than two observations. \textbf{(d)} The reference and source-varying models, both fitted on adult training speech. The source-varying model allows local rates to vary around the uncertain lexical and grammatical reference mean. \textbf{(e)} Gain in joint held-out log score per noun token for the source-varying model relative to the reference, evaluated across held-out families and across held-out corpora, for both speaker groups. \textbf{(f)} Illustration of the conditional randomization: the total number of articles within a stratum and each source\textquotesingle s noun count are held fixed while article occurrences are reallocated across sources. \textbf{(g)} The observed numbers of source cells with complete article absence exceed the null expectations: adult speech, 195 versus 170.23 (\(p=0.0005\)); child speech, 295 versus 273.12 (\(p=0.002\)). Histograms summarize 1,999 random reallocations; solid and dashed lines indicate observed counts and null means. Speaker artwork: Storyset/Freepik.}
\label{fig:6}
\end{minipage}%
}
\end{figure}

\subsection{4.7 Natural language use varies across sources after lexical
and syntactic
controls}\label{natural-language-use-varies-across-sources-after-lexical-and-syntactic-controls}

A premise of the hierarchical account is that even permitted forms can vary in their probability of use across communicative settings. We investigated whether such variation is detectable in natural language input. In child-directed and child speech, English nouns such as \emph{cake} may occur with or without articles. Figure~\ref{fig:6}a--c shows that article-use rates vary between families even when observations are grouped by the same word, grammatical number, and dependency role. We evaluated this pattern across eleven families from six CHILDES corpora, using the article indicator for \emph{a/an/the} as a measure of effective usage under these observed grammatical conditions.

If the variation reflects systematic differences between sources, a model that allows source-specific rates should better predict the distribution of observations in a new recording source. We compared a reference model that accounts for lexical and grammatical variation with an extension that also allows source rates to vary around the reference mean (Figure~\ref{fig:6}d). In leave-one-family-out evaluation, the source-varying model improved the joint marginal log score for adult speech in all eleven families (+0.0123 nats per noun token, token-weighted mean, 95\% CI \([0.0091,0.0155]\)) and for child speech in all eleven families (+0.0171 \([0.0107,0.0233]\); Figure~\ref{fig:6}e, left). Withholding complete corpora gave positive gains in all six recording collections: +0.0123 \([0.0098,0.0154]\) in adult speech and +0.0167 \([0.0109,0.0221]\) in child speech (Figure~\ref{fig:6}e, right). Supplement S9 also reports source-equal summaries. These comparisons show that modeling usage heterogeneity is useful for predicting held-out groups of observations.

We then asked whether the variation also appears as an excess of complete absences at the level of the observed family-by-word-by-grammar cells. For each word-by-number-by-role stratum, we randomly redistributed the observed total number of articles across families while keeping each family\textquotesingle s noun count fixed (Figure~\ref{fig:6}f). Under this homogeneous-use comparison, random sampling alone produces some cells with no articles. In adult speech, however, 195 of the 805 eligible source cells contained no articles, compared with a null expectation of 170.23 (95\% null range 158--183, \(p=0.0005\)). In child speech, 295 of 854 cells had no articles, compared with an expectation of 273.12 (258--287, \(p=0.0020\); Figure~\ref{fig:6}g). In both adult and child speech, article use also varied more across families within the same lexical and grammatical categories than expected under a common usage rate (Supplement S9). Naturalistic speech thus exhibits systematic clustering of article use across recording sources, even within comparable lexical and grammatical conditions. This structure is visible at the scale of families and entire recording collections, providing an empirical foundation for modeling source-dependent uncertainty about expected usage. Complete-absence probabilities provide a further way to examine the distribution of these usage rates, particularly its low-usage region. Supplement S10.1 investigates how well the models predict the probability of drawing groups of noun occurrences that contain no articles.

\subsection{4.8 Structural information and calibrated use rates draw on
different
cues}\label{structural-information-and-calibrated-use-rates-draw-on-different-cues}

\begin{figure}[!tp]
\makebox[\linewidth][c]{%
\begin{minipage}{158mm}
\captionsetup{font=integrated,skip=6pt}
\centering
\includegraphics[width=154mm,height=220mm,keepaspectratio]{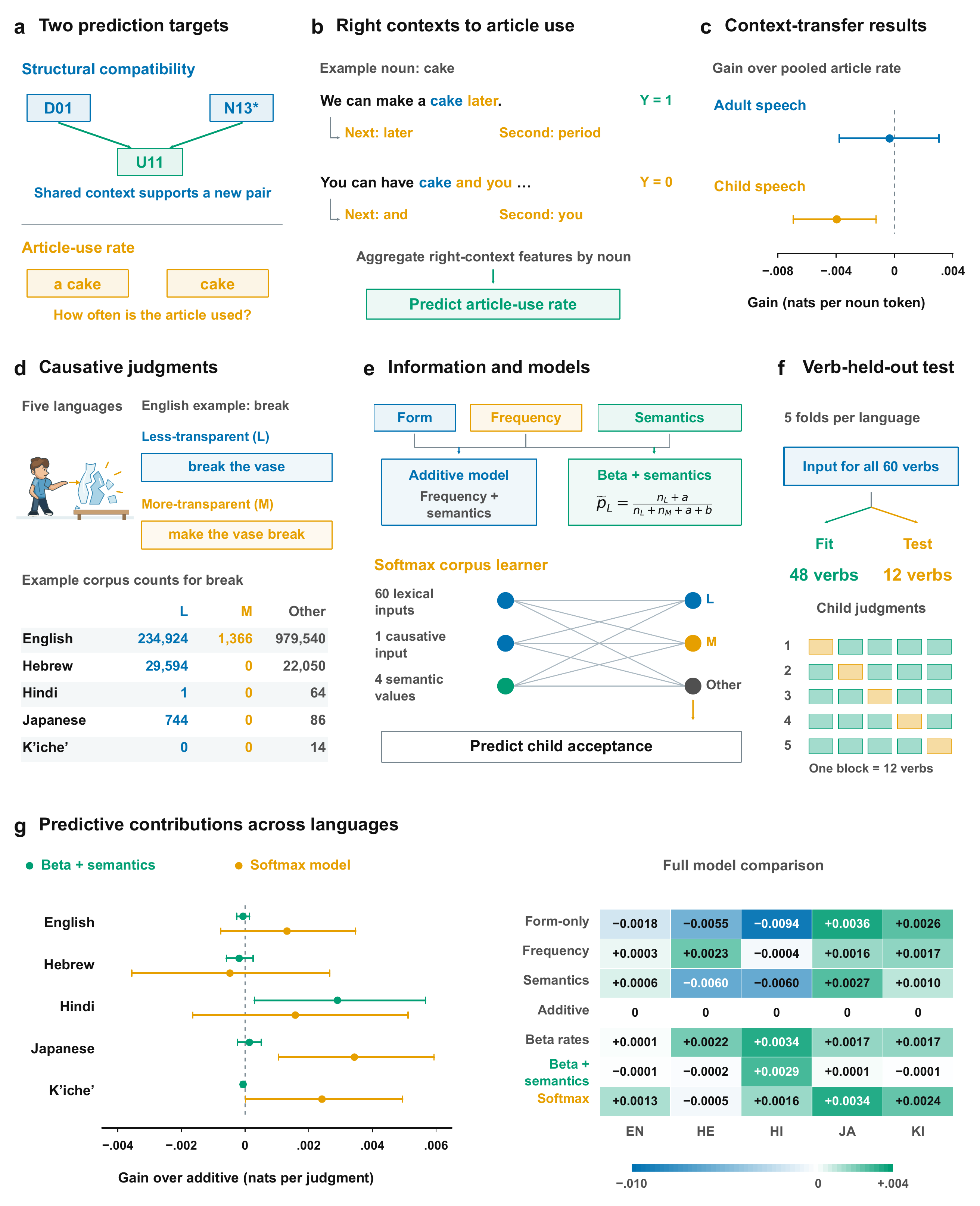}
\caption{\textbf{Distributional, lexical, and semantic information in structural generalization and form-use prediction.} \textbf{(a)} The two prediction targets: inferring whether a new word combination is structurally compatible, and estimating how often a form (here an article) is used. \textbf{(b)} The right-context features are aggregated by noun to predict article-use rates for held-out nouns. \textbf{(c)} Held-out gains over pooled article-use rates for adult and child speech, with 95\% word-bootstrap intervals. \textbf{(d)} The five-language causative dataset, with an English example of a less-transparent (L) and a more-transparent (M) expression of a causative event and the corresponding example lexical counts across languages. \textbf{(e)} The models use different combinations of form, frequency, and semantic information. The softmax corpus learner maps 65 inputs directly to competing L, M, and OTHER outputs, followed by the common model of behavioral judgments. \textbf{(f)} In each of five folds per language, the behavioral model is fitted to judgments for 48 verbs and evaluated on judgments for 12 held-out verbs. Corpus and semantic input remain available for all 60 verbs. \textbf{(g)} Gains in held-out judgment log score for the Beta-smoothed semantic and softmax models relative to the additive frequency-and-semantics model, with 95\% verb-bootstrap intervals. The accompanying matrix gives the complete seven-model comparison relative to the same reference. Scores give equal weight to verbs within each language.}
\label{fig:7}
\end{minipage}%
}
\end{figure}

The preceding analyses have examined two aspects of linguistic experience: shared continuations provide information about structural compatibility in controlled grammars (Section 4.1), and article-use rates vary systematically across natural recording sources (Section 4.7). We next asked whether the local distributional information associated with a noun can also predict its article-use rate (Figure~\ref{fig:7}a--b).

Using the first and second words following each noun as predictive features, the resulting model gives a mean gain over the pooled article-use rate of \(-0.0003\) nats per token in adult speech (95\% CI \([-0.0038,0.0031]\)) and \(-0.0040\) in child speech (\([-0.0069,-0.0013]\); Figure~\ref{fig:7}c). For both groups, the original continuation profiles scored below the mean of profiles permuted among nouns of similar frequency (Supplement S10.4). A representation based on each noun's distribution of syntactic dependency relations gives corresponding gains of \(-0.0019\) and \(-0.0032\) (Supplement S10.4). In these article-use prediction tasks, the pooled rates thus yield similar or better predictions than estimates transferred from short distributional or syntactic contexts. Together with the structural generalization results, this pattern highlights the importance of distinguishing the information used to infer compatibility from the information used to estimate actual usage. Lexical frequency, the meanings expressed in different situations, and competition among available forms offer further sources of information for the latter inference.

We therefore turned to a dataset in which children judged alternative causative forms associated with sixty verbs in each of English, Hebrew, Hindi, Japanese, and K'iche' \cite{art09}. The dataset provides lexical counts and semantic features that allow us to compare different ways of predicting judgments for held-out verbs (Figure~\ref{fig:7}d--f). In Hindi, Beta-smoothed lexical rates combined with semantic information improved on the additive raw-frequency-and-semantics model by 0.0029 nats per judgment (95\% CI \([0.0003,0.0057]\)). The corresponding gains in the other four languages were close to zero (\(-0.0002\) to +0.0001). A softmax corpus learner, which jointly uses lexical identity, causative marking, and semantic features to predict competing output forms, provided larger gains relative to the additive baseline in Japanese (+0.0034 \([0.0011,0.0059]\)) and K'iche' (+0.0024 \([0.00002,0.0050]\); Figure~\ref{fig:7}g). Relative to the Beta-smoothed model with semantic features, the softmax learner improved the mean log score by 0.0033 nats per judgment in Japanese and 0.0025 in K'iche'. The mean gains over the additive model were smaller in English (0.0013, 95\% CI \([-0.0008,0.0035]\)) and Hindi (0.0016, \([-0.0016,0.0051]\)). The complete seven-model comparison also reveals the contribution of general construction preferences. In Japanese and K'iche', the form-only model achieves slightly higher mean scores than the softmax learner (Supplement S11). Across the five languages, the predictive value of lexical smoothing, semantic information, and competition between forms varies with the linguistic system and the information available to each model. These analyses help characterize the cues through which learners can acquire expectations about linguistic forms. Scene-level observations of the kind described in Section 2.8 offer a complementary approach to studying how such expectations relate to semantic opportunities and conditional form choice.

\section{5. Discussion}\label{discussion}

\subsection{5.1 A generative connection between productivity and
restriction}\label{a-generative-connection-between-productivity-and-restriction}

At the beginning of the paper, we asked how learners can extend a grammatical pattern to new words while also learning the limits of that extension. Our account relates these abilities through the way expressions are generated. The linguistic contexts in which a word has appeared can provide evidence that a new combination is structurally possible. Experience with communicative situations can then inform how often the combination would be expected to occur if it were permitted. These expectations determine the evidential value of subsequent omissions. Structural permission, semantic opportunity, and form selection therefore contribute distinct information to a common inference problem.

The mathematical results show how this principle plays out when the learner is uncertain about the environment. The joint evidence law and context-refinement inequality explain why repeated experience in a familiar context and experience in new contexts can have different effects, even at the same observation count. Auxiliary calibration explains how positive experience elsewhere can change the learner\textquotesingle s interpretation of an unchanged target history. The reversal result shows how a stronger initial belief can eventually be outweighed by stronger evidence against it when the target was expected to occur more often. Finally, the identification analysis shows why observing communicative opportunities can help separate explanations that are equivalent when only the produced form is recorded. A common generative process gives rise to all of these predictions, linking the initial discovery of grammatical possibilities to the subsequent learning of their limits.

This account builds on hierarchical and sampling-based approaches to language learning. Perfors, Tenenbaum, and Wonnacott \cite{art10} show how learners can infer lexical and abstract constructional distributions simultaneously, including how variable construction use is across words. Related sampling effects occur in word learning \cite{art19}. Hsu and Griffiths \cite{art03} show that learners\textquotesingle{} interpretation of absence depends on what they believe generated the input: adults restricted an unattested combination more strongly when examples were presented as speaker productions (strong sampling) than when the same items were presented for classification with feedback (weak sampling). Our framework specifies the production probability under permission as \(q_t r_t\), distinguishing the chance of encountering a relevant meaning from the probability of selecting the target in that situation. This also relates naturally to the findings of Perek and Goldberg \cite{art17}, who showed that adults generalized according to construction meaning and were more conservative when a verb\textquotesingle s existing construction already served both event meanings. Such competition affects the conditional choice component \(r_{t}\) in our framework. By allowing uncertainty about the opportunities and resulting usage rates to persist within contexts and to be shared across them, we derive predictions about how the allocation and calibration of observations affect permission updates even when the total number of omissions is fixed.

\subsection{5.2 Why the organization of uncertainty
matters}\label{why-the-organization-of-uncertainty-matters}

The value of context diversity follows from the different explanations available to a learner. Repeated omissions within a single setting can share one explanation: the expression may have a low usage probability in that particular setting. As learners encounter more contexts with independently varying local rates, they accumulate information about a broader range of communicative environments. The possibility of a low usage tendency shared across those environments still provides a common explanation for their observations. The hierarchical model combines these two levels of uncertainty through the joint scale \(Cd_n\). The linear, logarithmic, and double-logarithmic growth regimes describe how the accumulation of evidence changes as learners face progressively richer sources of uncertainty about expected use.

The theory gives context diversity a specific statistical interpretation. Changes in conversational partner, topic, or communicative setting can provide new information when they expose the learner to different underlying usage rates. Proposition 2 formalizes this effect for contexts whose local rates are independently drawn conditional on a common shared mean. Our CHILDES analyses reveal systematic usage variation across families and recording collections, together with an excess of completely absent source cells relative to pooled lexical and grammatical rates. These findings establish source-dependent variation as an important property of naturalistic language use. Characterizing the dependence among usage rates across speakers, topics, and discourse states offers a way to determine how different forms of experience contribute to evidence accumulation.

Learners can also resolve uncertainty by observing other aspects of their environment. Frequent occurrences of related, permitted expressions may inform their expectations about target use, when those expressions sample comparable usage conditions. Alternatively, a learner may observe situations in which the target meaning is relevant and thereby refine expectations about opportunities themselves. In the latter case, the probability of selecting the target given an opportunity determines how much this knowledge increases expected use. Both routes predict that otherwise identical target omissions can have different effects after different calibration histories. This provides a way to manipulate the evidential value of absence without changing direct target exposure, using auxiliary expressions or independently presented scene cues.

\subsection{5.3 Structural-support reversal links positive and
restrictive
learning}\label{structural-support-reversal-links-positive-and-restrictive-learning}

The reversal prediction highlights a possible interaction between the two kinds of learning described above. Suppose a learner initially favors one unfamiliar expression and, because of independent communicative evidence, also expects it to be used relatively frequently if it is permitted. The learner may become increasingly surprised when opportunities to hear it repeatedly fail to produce it. If this difference in expected use is sufficiently large, the cumulative evidence can outweigh the initial structural advantage. In the fixed-rate model, the crossover is given by the initial log-odds advantage divided by the per-observation difference in omission evidence; in the hierarchical model, the relevant integrated likelihoods also depend on the allocation of contexts.

The reversal prediction concerns the trajectory through which learners acquire construction-specific restrictions. Two candidates may exchange their relative ordering during exposure as their initial structural advantages are gradually outweighed by differences in omission evidence. Tracking judgments throughout learning would reveal the emergence of the initial preference and the subsequent transition between the two candidates. An informative experiment would independently manipulate the amount of structural support and the expected use of candidate expressions, then track judgments after matched omission histories. Under the fixed-rate assumptions, Equation~\eqref{eq:11} gives a quantitative reference for when the ordering should reverse. The hierarchical account further predicts that redistributing the same number of observations across contexts can shift that point. Such a design would directly test the proposed interaction between productive generalization and restrictive learning.

\subsection{5.4 Linking computational mechanisms to behavioral and naturalistic evidence}\label{behavioral-and-natural-input-findings-specify-the-learning-interfaces}

The analyses in this paper address complementary parts of the learning problem. The controlled grammars illustrate how a learner could obtain structural support without direct experience of a combination. The randomized conditions in Samara et al.\textquotesingle s experiments establish the importance of meaning-matched competition, and our paired reanalysis shows that information specific to restricted items helps predict production choices. The naturalistic analyses show that a model of expression use benefits from allowing source-dependent usage rates. Finally, the word-held-out transfer and cross-linguistic comparisons examine which distributional, lexical, and semantic cues contribute to different predictions. These findings provide empirical and computational grounding for the proposed components, and the precise context-allocation and calibration predictions suggest how they can be investigated together.

The distinction between structural compatibility and expected usage helps explain the different contributions of distributional information across our analyses. Shared continuations provide structural evidence in the artificial grammars, while estimating article-use rates for held-out nouns involves additional information about the meanings, grammatical environments, and pragmatic circumstances in which those nouns occur. The cross-linguistic comparisons further illustrate the contributions of lexical experience and competing forms to judgments of linguistic acceptability. These findings suggest a computational division of labor in which different aspects of experience contribute to the discovery of grammatical patterns and the estimation of how those patterns are used.

Accurate estimation of usage also involves characterizing the low-rate region of the distribution, which plays a central role in the theory of prolonged absence. Supplement S10.1 examines this issue by comparing observed and predicted probabilities of drawing article-free subsets of noun occurrences. In child speech, the observed probability for a random subset of sixteen tokens is 0.054, while the source-varying Beta model predicts 0.141. This pattern reveals a particular challenge for modeling rare complete-absence events: the model must accurately represent the environments in which the form is used infrequently. Incorporating developmental changes, pragmatic definiteness, topic persistence, and finer-grained discourse dynamics offers promising avenues for characterizing these environments and extending the source-level analysis to shorter communicative timescales.

The paired behavioral analysis also suggests how a generative model of language knowledge could be linked to different forms of experimental measurement. A learner may possess knowledge of a particular word\textquotesingle s restrictions that influences both their acceptability judgments and their own productions. These behaviors express that knowledge through different response processes. In our analysis, a model accounting for condition-specific ordinal thresholds and local restricted-item effects captured most of the judgment information, while restricted ratings provided additional information about active production. A joint measurement model can connect these response processes to a common learning state containing structural permission and conditional choice preferences. Estimating both components from coordinated judgment and production tasks would help characterize how grammatical knowledge and expectations about form use contribute to each kind of behavior.

\subsection{5.5 Observation designs for testing counterfactual
opportunity
inference}\label{observation-designs-for-testing-counterfactual-opportunity-inference}

A central experimental challenge is to determine whether learners evaluate an absent expression in relation to the opportunities in which it could have been used. Ordinary language records usually tell us which expressions were produced. Scene-level records can also reveal whether the relevant meaning was available when those expressions were produced. The Fisher information calculation shows that these two measurement designs constrain different parameters: target-only observations estimate qr, whereas observing opportunities and choices within them separately identifies \(q\) and \(r\) conditional on permission. A full permission-learning experiment would additionally provide structural cues or known licensing conditions to inform \(Z\). Such a design would allow the theoretical effects of context allocation and opportunity calibration to be manipulated within a common cohort.

Three comparisons follow from the theory. First, learners could receive the same total number of target omissions, distributed either within a persistent context or across contexts designed to provide independently varying local rates. The context-refinement result predicts greater restriction in the latter condition. Second, learners with the same target histories could receive different positive calibration experiences, either through matched auxiliary expressions or through labeled scenes. Proposition 3 predicts that a history indicating more frequent relevant use or opportunities will make subsequent omissions more informative. Third, initial structural support and expected usage could be manipulated independently, with learners\textquotesingle{} judgments measured repeatedly as omissions accumulate. The model predicts a relative preference reversal when the difference in cumulative absence evidence exceeds the initial structural advantage. These designs extend the artificial-language paradigms exemplified by Samara et al. \cite{art02} by varying context allocation and calibration at the same total number of training exposures, and provide direct quantitative tests of the mechanisms derived here.

The feasibility of distinguishing the proposed generating mechanisms can be examined through the model-recovery simulation in Supplement S12. When the candidate model parameters are fixed, the probabilities of recovering the known fixed-rate, persistent local-rate, and uncertain shared-mean mechanisms from complete-absence outcomes reach 98.6\%, 97.3\%, and 100.0\%, respectively, with 450 independent lexical items. These results provide a quantitative reference for designing experiments that distinguish different structures of usage-rate uncertainty through the allocation of observations across contexts.

These predictions combine general properties of omission inference with properties of a particular parametric model. Beta distributions make the hierarchical updates tractable and yield the explicit joint scaling law. The finite-sample context-refinement inequality applies to arbitrary conditional local-rate distributions on {[}0,1{]}, while the asymptotic form depends on the probability mass the model assigns to rates close to zero (Supplement S2). This distinction suggests both a statistical and a cognitive research program: use richer multimodal recordings to characterize opportunities and low-rate environments, and develop joint behavioral measurements to test how learners use this information when they judge and produce language. In this way, the inference from an expression\textquotesingle s possible use to the significance of its actual absence can be examined more directly.

\section{6. Conclusion}\label{conclusion}

Language learners discover both the possibilities of a grammatical pattern and the limits of those possibilities. Our framework connects these two aspects of learning by distinguishing whether a form is permitted, whether its meaning is relevant, and whether a speaker selects it. Shared predictive continuations provide a route to initial structural support, and the subsequent evidential value of an omission depends on the expected use of the expression. The mathematical results describe how this evidence changes with the distribution of observations across contexts, with positive experience that calibrates usage, and with the relative strength of initial structural support and expected use. Observing communicative opportunities also makes it possible to distinguish the availability of a meaning from the speaker's selection of a particular expression. The artificial grammars reveal how shared predictions can support new combinations, the behavioral analyses reveal how meaning-matched competition shapes linguistic restriction, and the naturalistic analyses establish the importance of variation in expected usage across communicative sources. Together, these findings develop a common principle of language learning: what remains unsaid becomes informative through a learner's knowledge of when it could have been said. Through this principle, productive generalization and the acquisition of construction-specific restrictions become interconnected aspects of the same learning process.

\section{Data and code}\label{data-and-code}

The behavioral reanalyses use the artificial-language data of Samara et al. \cite{art20} and the five-language causative-learning data of Ambridge et al. \cite{art21}. The natural-language analyses use the gold annotations in UD-English-CHILDES \cite{art22}. Supplement S12 identifies the original data sources. Analysis scripts and derived results are available from the corresponding author.

\subsection*{Funding and competing interests}

This research received no specific funding. The author declares no competing interests.

\subsection*{Ethics}

The study uses simulated data and reanalyzes existing behavioral and corpus records. The original research articles and corpus documentation describe the collection of the archived human observations.

\subsection*{Artwork credit}

Figures 2--6 use adaptations of \href{https://storyset.com/illustration/coffee-shop/pana}{Coffee shop (Pana) by Storyset/Freepik}. The illustrations are credited under the Storyset terms.

\FloatBarrier


\section{References}\label{references}

\begin{thebibliography}{99}
\setlength{\itemsep}{1pt plus 1pt}
\bibitem{art01}
Reeder, P. A., Newport, E. L., \& Aslin, R. N. (2013). From
shared contexts to syntactic categories: The role of distributional
information in learning linguistic form-classes. \emph{Cognitive
Psychology}, 66, 30--54. DOI: 10.1016/j.cogpsych.2012.09.001.

\bibitem{art02}
Samara, A., Wonnacott, E., Saxena, G., Maitreyee, R., Fazekas,
J., \& Ambridge, B. (2025). Learners restrict their linguistic
generalizations using preemption but not entrenchment: Evidence from
artificial-language-learning studies with adults and children.
\emph{Psychological Review}, 132, 1--17. DOI: 10.1037/rev0000463.

\bibitem{art03}
Hsu, A. S., \& Griffiths, T. L. (2016). Sampling assumptions affect use of indirect negative evidence in language learning. \emph{PLOS ONE}, 11, e0156597. DOI: 10.1371/journal.pone.0156597.

\bibitem{art04}
Hsu, A. S., Horng, A., Griffiths, T. L., \& Chater, N. (2017;
online 2016). When absence of evidence is evidence of absence: Rational
inferences from absent data. \emph{Cognitive Science}. DOI:
10.1111/cogs.12356.

\bibitem{art08}
Yang, X., Ju, Z., Bu, L., Liu, Z., \& Schneider, N. (2025).
UD-English-CHILDES: A collected resource of gold and silver Universal
Dependencies trees for child language interactions. \emph{Proceedings of
UDW / SyntaxFest 2025}, 52--58.

\bibitem{art09}
Ambridge, B., Doherty, L., Maitreyee, R., Tatsumi, T.,
Zicherman, S., et al.~(2022). Testing a computational model of causative
overgeneralizations: Child judgment and production data from English,
Hebrew, Hindi, Japanese and K'iche'. \emph{Open Research Europe}. DOI:
10.12688/openreseurope.13008.2. Data: https://osf.io/7f2dg/.

\bibitem{art10}
Perfors, A., Tenenbaum, J. B., \& Wonnacott, E. (2010).
Variability, negative evidence, and the acquisition of verb argument
constructions. \emph{Journal of Child Language}, 37, 607--642. DOI:
10.1017/S0305000910000012.

\bibitem{art11}
Mintz, T. H. (2003). Frequent frames as a cue for grammatical
categories in child directed speech. \emph{Cognition}, 90, 91--117.
DOI: 10.1016/S0010-0277(03)00140-9.

\bibitem{art12}
Wonnacott, E., Newport, E. L., \& Tanenhaus, M. K. (2008).
Acquiring and processing verb argument structure: Distributional learning
in a miniature language. \emph{Cognitive Psychology}, 56, 165--209.
DOI: 10.1016/j.cogpsych.2007.04.002.

\bibitem{art13}
Wonnacott, E. (2011). Balancing generalization and lexical
conservatism: An artificial language study with child learners.
\emph{Journal of Memory and Language}, 65, 1--14.
DOI: 10.1016/j.jml.2011.03.001.

\bibitem{art14}
Brooks, P. J., \& Tomasello, M. (1999). How children constrain
their argument structure constructions. \emph{Language}, 75, 720--738.
DOI: 10.2307/417731.

\bibitem{art15}
Boyd, J. K., \& Goldberg, A. E. (2011). Learning what not to say:
The role of statistical preemption and categorization in a-adjective
production. \emph{Language}, 87, 55--83. DOI: 10.1353/lan.2011.0012.

\bibitem{art16}
Goldberg, A. E. (2011). Corpus evidence of the viability of
statistical preemption. \emph{Cognitive Linguistics}, 22, 131--153.
DOI: 10.1515/COGL.2011.006.

\bibitem{art17}
Perek, F., \& Goldberg, A. E. (2017). Linguistic generalization
on the basis of function and constraints on the basis of statistical
preemption. \emph{Cognition}, 168, 276--293.
DOI: 10.1016/j.cognition.2017.06.019.

\bibitem{art18}
Ambridge, B., Bidgood, A., Twomey, K. E., Pine, J. M.,
Rowland, C. F., \& Freudenthal, D. (2015). Preemption versus entrenchment:
Towards a construction-general solution to the problem of the retreat
from verb argument structure overgeneralization. \emph{PLOS ONE}, 10,
e0123723. DOI: 10.1371/journal.pone.0123723.

\bibitem{art19}
Xu, F., \& Tenenbaum, J. B. (2007). Sensitivity to sampling in
Bayesian word learning. \emph{Developmental Science}, 10, 288--297.
DOI: 10.1111/j.1467-7687.2007.00590.x.

\bibitem{art20}
Samara, A., Wonnacott, E., Saxena, G., Maitreyee, R., Fazekas,
J., \& Ambridge, B. (2025). Data for the artificial-language-learning
studies [Data set]. OSF. \url{https://osf.io/4jdvf/}

\bibitem{art21}
Ambridge, B., et al. (2022). Data for the five-language
causative-learning study [Data set]. OSF. \url{https://osf.io/7f2dg/}

\bibitem{art22}
Yang, X., Ju, Z., Bu, L., Liu, Z., \& Schneider, N. (2025).
UD-English-CHILDES [Data set]. Universal Dependencies.
\url{https://github.com/UniversalDependencies/UD_English-CHILDES}
\end{thebibliography}
\end{document}


\maketitle

\setstretch{1.12}

The main article distinguishes three components of language use: structural permission, the availability of semantic opportunities, and the selection of an expression. Sections S1--S6 develop the probabilistic model underlying these components, derive the evidence accumulation laws, and establish the observation conditions under which the relevant parameters are identifiable. Sections S7--S11 describe the computational, behavioral, and naturalistic analyses that examine different aspects of the framework. Section S12 investigates how complete-absence observations can distinguish the proposed generating mechanisms and provides a quantitative reference for experimental design.

\section{S1. Structural permission and the likelihood of
absence}\label{s1.-structural-permission-and-the-likelihood-of-absence}

Bayesian accounts of language learning have shown that the value of indirect negative evidence depends on how a learner expects utterances to be sampled \cite{si03}, on how strongly the missing form was expected \cite{si04}, and on how evidence is combined across levels of constructional abstraction \cite{si10}. Here we specify the joint model used to combine initial structural evidence with later omissions, and extend the strict prohibition model to allow production errors. We use a shared effective-use mean \(\mu\) to combine evidence across communicative contexts.

Let \(S\) denote structural evidence and \(P(Z,\mu \mid S)\) the learner\textquotesingle s joint distribution over the target\textquotesingle s permission \(Z\) and the shared mean \(\mu\) after observing \(S\). In a factorized specification, \(Z\mid S\sim \operatorname{Bernoulli}(\pi (S))\) and \(\mu \mid S\sim \operatorname{Beta}(A(S),B(S))\) are independent conditional on \(S\). Structural evidence may thus inform the probability of permission and the expected usage rate through different parameters. Given \(\mu\), each context has an independently drawn local rate. Within a context, observations share that rate; conditional on the local rates and \(Z\), the individual emissions are independent.

On occasion \(t\), let \(q_{t}=P(O_{t}=1)\) denote the probability that the relevant semantic opportunity is present (with the conditioning on the setting and structural evidence suppressed). Conditional on permission and that opportunity, the target is chosen with probability \(r_{t}=P(Y_{t}=1\mid O_{t}=1,Z=1)\). Under the strict emission model, a target occurrence requires all three conditions. Its effective probability under permission is therefore \(\theta _{t}=q_{t} r_{t}\). When an alternative form is used in a labeled opportunity, the observation constrains the conditional choice probability. When only an omission is recorded, the likelihood instead averages over both the possible availability of the meaning and the possible choices of the speaker.

Let \(D_{0}\) denote a history containing no target occurrences, and \(L(D_{0}\mid S)\) its likelihood conditional on permission. Under strict prohibition, \(D_{0}\) has probability one. Applying Bayes\textquotesingle{} rule gives

\[P(Z=1\mid D_0,S)=\frac{\pi(S)L(D_0\mid S)}{1-\pi(S)+\pi(S)L(D_0\mid S)}.\]

The two terms in the denominator correspond to the prohibited and permitted explanations of the same omission history. If prohibited forms can occasionally be produced as errors, the likelihood of the history under prohibition becomes \(L_{0}=\mathbb{E}[\prod _{t}(1-e_{t})]\), where \(e_{t}\) is the effective error rate on occasion \(t\). Writing \(L_{1}\) for the likelihood under permission, the log-odds update is then \(\log  L_{1}-\log  L_{0}\). This compares the probability of the observed omissions under both explanations. Under the strict permission model, an occasion with no semantic opportunity has \(\theta _{t}=0\) and contributes a factor of one to the omission likelihood.

The effective-rate hierarchy used in the main article places a distribution directly on \(\theta =qr\) through the shared mean \(\mu\). Supplement S4 develops a second formulation in which scene labels inform a separate shared mean \(\nu\) for semantic opportunity \(q\), and Supplement S6 establishes the information required to identify \(q\) and \(r\) separately.

\section{S2. Why the joint evidence law is governed by low use
rates}\label{s2.-why-the-joint-evidence-law-is-governed-by-low-use-rates}

Long histories containing no occurrences of a target are difficult to explain under models that assign it a substantial probability of use. As the history grows, the likelihood under the shared-mean hierarchy is increasingly concentrated near \(\mu =0\). The asymptotic behavior of the integral is therefore determined by the size of the low-rate region that can still explain the omissions and by the prior probability assigned to that region. We make this argument uniform in both the number of contexts \(C\) and the number of observations per context \(n\).

Fix \(A,B,\kappa >0\) and define

\[c_j=\frac{\kappa}{\kappa+j},\quad d_n=\sum_{j=0}^{n-1}c_j,\quad m_n(\mu)=\prod_{j=0}^{n-1}(1-c_j\mu).\]

The factor \(m_{n}(\mu )\) is the conditional probability of \(n\) omissions within one persistent context. Since the local rates are independent given \(\mu\), \(C\) contexts contribute \(m_{n}(\mu )^C\). Integrating with respect to the Beta prior on the shared mean gives

\[L_{C,n}=\frac{1}{B(A,B)}\int_0^1\mu^{A-1}(1-\mu)^{B-1}m_n(\mu)^C\,d\mu.\]

\textbf{Joint evidence law.} For arbitrary integer sequences \(C,n\geq 1\) with \(T=Cd_{n}\to \infty\),

\[L_{C,n}\sim\frac{\Gamma(A+B)}{\Gamma(B)}T^{-A}.\]

\textbf{Proof.} We first approximate the omission factor near the low-rate boundary. Fix \(0<\delta <1\). The expansion of \(\log (1-x)\) gives, uniformly in \(n\) and for \(0\leq \mu \leq \delta\),

\[\log m_n(\mu)=-d_n\mu+R_n(\mu),\qquad |R_n(\mu)|\leq K_\delta\mu^2\sum_{j=0}^{\infty}c_j^2.\]

The sum of squared coefficients is finite, since \(c_{j}\) decreases as \(1/j\) for large \(j\). The leading term for \(C\) contexts is \(-Cd_{n} \mu =-T\mu\). We accordingly rescale the boundary region by \(t=T\mu\). For any fixed \(t\), the error in \(C \log  m_{n}(t/T)\) is bounded by a constant times \(t^{2}/(Cd_{n}^{2})\). Since \(d_{n}\geq 1\) and \(Cd_{n}\to \infty\), this bound tends to zero, leaving the limiting exponent \(-t\).

To justify integrating this limit, use the inequality \(m_{n}(\mu )^C\leq \exp (-T\mu )\). On the rescaled region \(0\leq t\leq T\delta\), the factor \((1-t/T)^{B-1}\) remains bounded for every \(B>0\). After rescaling, the integrand is bounded by a constant times \(t^{A-1}\exp (-t)\), an integrable function for \(A>0\). Dominated convergence therefore yields

\[T^A\int_0^\delta\mu^{A-1}(1-\mu)^{B-1}m_n(\mu)^C\,d\mu\longrightarrow\Gamma(A).\]

It remains to control the portion of the integral away from zero. For \(\mu \in [\delta ,1]\), the omission factor is bounded above by \(\exp (-T\delta )\), so this portion is at most \(\exp (-T\delta )\) times a finite Beta integral. It vanishes even after multiplication by \(T^A\). Dividing the limiting boundary integral by \(B(A,B)\) gives the constant \(\Gamma (A+B)/\Gamma (B)\), establishing the stated asymptotic expression. \(\square\)

The scale \(T\) has a simple interpretation. The digamma identity \(d_{n}=\kappa [\psi (\kappa +n)-\psi (\kappa )]=\kappa  \log  n+O(1)\) \cite{si11} shows that adding observations within one context increases \(d_{n}\) with decreasing marginal weights. Increasing \(C\) multiplies this accumulated weight by the number of independently drawn local contexts. Taking \(-\log\) of the asymptotic likelihood yields the following regimes. When \(n=1\), the exact identity \(m_{1}(\mu )=1-\mu\) also provides a simple finite-sample expression.

\Needspace{12\baselineskip}

\textbf{Asymptotic evidence-growth regimes for the shared-mean hierarchy and the exact likelihood when each context supplies one observation.}

{\def\LTcaptype{none}
\begin{longtable}[]{@{}ll@{}}
\toprule\noalign{}
Experience structure in the shared-mean hierarchy & Evidence growth or exact likelihood \\
\midrule\noalign{}
\endhead
\bottomrule\noalign{}
\endlastfoot
Fixed positive \(n\), increasing \(C\) & \(A\log C+O(1)\) \\
Fixed positive \(C\), increasing \(n\) & \(A\log\log n+O(1)\) \\
Jointly increasing \(C,n\) & \(A\log(Cd_n)+O(1)\) \\
One observation per context & Exact likelihood \(B(A,B+C)/B(A,B)\) \\
\end{longtable}
}

These rates of accumulation depend on how easily the generating model can explain the absence through very low usage probabilities. If every rate is bounded below by \(\theta _{\mathrm{min}}>0\), an omission sequence of length \(n\) has likelihood at most \((1-\theta _{\mathrm{min}})^n\), and evidence grows at least as fast as \(-n \log (1-\theta _{\mathrm{min}})\). Conversely, positive probability mass on a zero-use state can give a positive limiting omission likelihood when that state persists over the relevant history. In the continuous Beta hierarchy, it is the prior density proportional to \(\mu ^{A-1}\) near zero that produces the power A in the joint asymptotic law. Figure 2f compares exact likelihoods with this law across 75 combinations of hyperparameters, context counts, and within-context durations; numerical quadrature uses a rescaling near \(\mu =0\) to resolve the boundary region.

\section{S3. How splitting a context changes the
evidence}\label{s3.-how-splitting-a-context-changes-the-evidence}

The asymptotic result in S2 describes the accumulation of evidence when observations become numerous. For an experiment with a fixed total number of observations, we can obtain a distribution-independent comparison by changing whether two blocks of omissions share the same local rate.

Suppose \(a+b\) omissions arise in a single persistent context. Conditional on \(\mu\), let \(X=1-\theta \in [0,1]\) be the probability of a target omission and let \(X'\) be an independent copy drawn from the same conditional distribution. Both \(X^a\) and \(X^b\) are increasing functions of \(X\). Consequently,

\[\operatorname{Cov}(X^a,X^b)=\frac12\mathbb E[(X^a-X'^a)(X^b-X'^b)]\geq0.\]

The product of the two differences is nonnegative for every pair of draws, because the functions vary in the same direction. This gives

\[m_{a+b}(\mu)\ge m_a(\mu)m_b(\mu),\]

with strict inequality when the conditional local-rate distribution is nondegenerate. The first side is the likelihood under a shared persistent local rate; the second is the likelihood when the two blocks have independently sampled local rates conditional on \(\mu\). Multiplication by other context factors and integration over the same shared mean preserve this ordering. Applying the argument successively to every split establishes the result for any refinement of the context partition.

As in work on sampling and generalization in word learning \cite{si12}, the allocation of the same number of observations changes how they constrain possible generating processes. In a single context, one low local rate accounts for both blocks of omissions. In separate, conditionally independent contexts, the likelihood averages over distinct rates. The latter history is less probable under permission and thus carries stronger evidence against it. This is an exact comparison at fixed total exposure and depends on the generative independence of the refined contexts. Figure 2d illustrates it; all 140 parameter and allocation regimes examined numerically obeyed the ordering.

\section{S4. How positive experience calibrates later
omissions}\label{s4.-how-positive-experience-calibrates-later-omissions}

The evidence carried by target omissions depends on the learner's expectations about how often the target would occur under permission. These expectations can develop through experience with other expressions and with the communicative environment itself. We derive two corresponding mechanisms of calibration: observations of related expressions that share the target's effective-use environment, and observations of scenes that provide information specifically about semantic opportunities.

\subsection{S4.1 Observations of related
expressions}\label{s4.1-observations-of-related-expressions}

Suppose that the learner encounters expressions that are known to be permitted and whose usage rates are drawn from the same conditional environment as the target. If each auxiliary probe samples a separate context independently conditional on \(\mu\), then a binary indicator \(X_{j}\) recording whether the auxiliary expression is used has marginal probability \(\mu\). We therefore model the auxiliary probes as independent \(X_{j}\mid \mu \sim \operatorname{Bernoulli}(\mu )\). With a \(\operatorname{Beta}(A,B)\) prior and \(k\) positive outcomes in \(M\) probes, Bayes\textquotesingle{} rule gives \(\mu \mid X\sim \operatorname{Beta}(A+k,B+M-k)\). In the factorized model used for Table~\ref{tab:S1}, target permission is independent of the shared mean given structural evidence, so the auxiliary history leaves its initial permission at 0.8.

Let \(n\) be an allocation of later target omissions and write their conditional likelihood as \(G(\mu )=\prod _{c} m_{n_c}(\mu )\). Each factor decreases with \(\mu\): an expression with a higher expected usage rate is less likely to remain absent. For fixed \(M\), increasing \(k\) by one changes the posterior density of \(\mu\) by a multiplicative factor proportional to \(\mu /(1-\mu )\). This factor is increasing in \(\mu\), so the posteriors are ordered by monotone likelihood ratio \cite{si13}. More positive probes shift \(\mu\) upward in stochastic order and decrease the posterior expectation \(\mathbb{E}[G(\mu )]\). It follows that a learner with more positive auxiliary experience assigns a lower likelihood to the same target omissions under permission.

The auxiliary observations also alter the long-run evidence law. For fixed \(M\) and \(k\), applying the result of S2 to the updated Beta parameters gives

\[L_{C,n}\sim\frac{\Gamma(A+B+M)}{\Gamma(B+M-k)}(Cd_n)^{-(A+k)}.\]

Consequently, \(I_{C,n}=(A+k)\log (Cd_{n})+O(1)\). The coefficient of evidence accumulation on the logarithmic joint scale increases from A to \(A+k\). Positive experience thus constrains the low shared-rate explanation that would otherwise remain available during prolonged absence.

\subsection{S4.2 Scene labels and conditional
choice}\label{s4.2-scene-labels-and-conditional-choice}

An observation of a related expression combines information about its meaning and about the form selected by the speaker. A second design supplies information about the meaning directly. Here let \(\nu \sim \operatorname{Beta}(A,B)\) be the shared probability that a relevant opportunity is available, and let \(q_{c}\mid \nu \sim \operatorname{Beta}(\kappa \nu ,\kappa (1-\nu ))\). Independent scene labels \(O_{j}\mid \nu \sim \operatorname{Bernoulli}(\nu )\) update the shared opportunity mean. A permitted target has a known conditional choice probability \(r_{0}\in (0,1]\), so the local effective rate is \(r_{0}q_{c}\). The conditional likelihood of \(n\) omissions in a context is

\[h_n(\nu;r_0)=\mathbb E[(1-r_0q_c)^n\mid\nu]=\sum_{j=0}^n{n\choose j}(1-r_0)^{n-j}r_0^j m_j(\nu),\]

with \(m_{0}=1\). The second expression follows by expanding \((1-r_{0}+r_{0}(1-q_{c}))^n\) and integrating the powers of \(1-q_{c}\). At fixed \(\kappa\), the conditional Beta distribution of \(q_{c}\) shifts upward in stochastic order as \(\nu\) increases. Since the omission probability decreases with \(q_{c}\), \(h_{n}\) decreases with \(\nu\). Averaging \(\prod_c h_{n_c}(\nu;r_0)\) over the scene-label posterior gives the same qualitative calibration effect as in S4.1.

When \(r_{0}=1\), every opportunity leads to target selection under permission, so \(h_{n}=m_{n}\). When \(r_{0}\) is smaller, a target can remain absent even in a relevant scene, and the effective rate lies in \([0,r_{0}]\). To illustrate the role of this distinction, Table~\ref{tab:S1} holds \(A=1\), \(B=4\), \(\kappa =4\), \(M=12\), \(C=4\), \(n=4\), and initial permission 0.8 fixed. For each choice rate, the table reports the likelihood of the same sixteen target omissions after different numbers of positive probes, and the resulting posterior permission probability.

\begin{longtable}[]{@{}
  >{\raggedleft\arraybackslash}p{(\linewidth - 6\tabcolsep) * \real{0.1852}}
  >{\raggedleft\arraybackslash}p{(\linewidth - 6\tabcolsep) * \real{0.2346}}
  >{\raggedleft\arraybackslash}p{(\linewidth - 6\tabcolsep) * \real{0.2840}}
  >{\raggedleft\arraybackslash}p{(\linewidth - 6\tabcolsep) * \real{0.2963}}@{}}
\caption{Effects of auxiliary calibration on the likelihood of a fixed target omission history and the resulting posterior permission probability, for effective-rate and scene-label observation models.\label{tab:S1}}\tabularnewline
\toprule\noalign{}
\begin{minipage}[b]{\linewidth}\raggedleft
Choice rate
\end{minipage} & \begin{minipage}[b]{\linewidth}\raggedleft
Positive probes
\end{minipage} & \begin{minipage}[b]{\linewidth}\raggedleft
Omission likelihood
\end{minipage} & \begin{minipage}[b]{\linewidth}\raggedleft
Posterior permission
\end{minipage} \\
\midrule\noalign{}
\endfirsthead
\toprule\noalign{}
\begin{minipage}[b]{\linewidth}\raggedleft
Choice rate
\end{minipage} & \begin{minipage}[b]{\linewidth}\raggedleft
Positive probes
\end{minipage} & \begin{minipage}[b]{\linewidth}\raggedleft
Omission likelihood
\end{minipage} & \begin{minipage}[b]{\linewidth}\raggedleft
Posterior permission
\end{minipage} \\
\midrule\noalign{}
\endhead
\bottomrule\noalign{}
\endlastfoot
1.0 & 0 & 0.5702 & 0.6952 \\
1.0 & 3 & 0.0901 & 0.2650 \\
1.0 & 6 & 0.0104 & 0.0401 \\
0.6 & 0 & 0.6677 & 0.7276 \\
0.6 & 3 & 0.1792 & 0.4175 \\
0.6 & 6 & 0.0393 & 0.1359 \\
\end{longtable}

The lower choice rate \(r_{0}=0.6\) softens the update relative to \(r_{0}=1\): even if the relevant meaning is common, speakers may prefer an alternative. In both observation models, additional positive probes make subsequent target absence less probable under permission. Table~\ref{tab:S1} assumes auxiliary probes from independent contexts. Multiple probes drawn from the same persistent context would instead require averaging their joint likelihood over the common local rate, just as for the target observations.

\section{S5. When initially stronger support
reverses}\label{s5.-when-initially-stronger-support-reverses}

The structural prior establishes the learner's initial ordering of candidate expressions. Expectations about their usage rates then determine how the subsequent omissions change that ordering. To describe how these contributions interact, consider a candidate \(i\) with a known effective rate \(\theta _{i}\). Its posterior log odds after \(n\) omissions are \(\operatorname{logit}\pi_i + n\log(1-\theta_i)\). Comparing two candidates \(H\) and \(L\) therefore gives an initial difference in prior log odds plus a term linear in \(n\). When \(\pi _{H}>\pi _{L}\) and \(\theta _{H}>\theta _{L}\), the first term favors \(H\) and the slope of the omission term is negative. Setting the difference to zero gives \(n\)* in Equation (11) of the main text; the first strict reversal occurs at \(\lfloor n*\rfloor +1\).

For uncertain rates, the evidence contributed by the omission history is computed by integrating over the candidate\textquotesingle s distribution of possible rates. In our hierarchical illustration, \(\pi _{H}=0.9\) and \(\pi _{L}=0.6\), with \(\mu _{H}\sim \operatorname{Beta}(4,8)\) and \(\mu _{L}\sim \operatorname{Beta}(1,11.5)\); both have \(\kappa =4\). Each candidate\textquotesingle s likelihood is evaluated using its own shared-rate distribution. With four contexts and two observations per context, the posterior permission difference \(H-L\) is approximately +0.0064. With four observations per context, it is \(-0.1662\). Figure 3f varies context number through \(C=1,2,4,8,16,32\) and the within-context counts through 64. The surface shows that the ordering depends on the context allocation and repetition, as well as on the initial probabilities and usage priors.

Perek and Goldberg \cite{si14} found that construction function supports generalization while a competing form constrains it. Our reversal prediction suggests a way to examine their interaction over time. By independently manipulating structural support and calibrated expectations about use, and then presenting matched omission histories, a future experiment could measure whether learners\textquotesingle{} preferences follow the predicted crossover.

\section{S6. What opportunity labels
identify}\label{s6.-what-opportunity-labels-identify}

When a target expression is absent, the learner or researcher may have observed a situation in which it was irrelevant, or a relevant situation in which a different form was chosen. To quantify what information is gained by distinguishing these cases, we compare the likelihoods of occurrence-only data with data that also record the opportunity. Throughout this calculation we condition on \(Z=1\).

Suppose \(y\) target occurrences are observed in \(N\) independent occasions with constant \(q\) and \(r\). If only \(Y\) is recorded, the log likelihood is

\[\ell_Y(q,r)=y\log(qr)+(N-y)\log(1-qr).\]

All parameter pairs with the same product give the same likelihood. The gradient of qr is (r,q), so the Fisher information matrix in Equation (12) is proportional to the outer product of this vector with itself and has rank one.

If \(O\) is also observed, the data distinguish three events: no relevant opportunity, an opportunity followed by an alternative form, and an opportunity followed by the target. Their probabilities are

\[P(O=0,Y=0)=1-q,\quad P(O=1,Y=0)=q(1-r),\quad P(O=1,Y=1)=qr.\]

Let \(N_{O}\) denote the number of relevant opportunities and \(N_{Y}\) the number of target selections. The log likelihood then separates into \(N_{O} \log  q+(N-N_{O})\log (1-q)\), which concerns the presence of opportunities, and \(N_{Y} \log  r+(N_{O}-N_{Y})\log (1-r)\), which concerns selections within those opportunities. The expected cross-information is zero. The diagonal information terms are \(N/[q(1-q)]\) and \(Nq/[r(1-r)]\), both positive at interior parameter values. By the standard relation between information rank and local identification in regular parametric models \cite{si15}, the two rates are separately identifiable; the occurrence-only likelihood identifies their product instead. Known conditional choice rates or interventions with specified likelihoods can provide related identifying information.

Conditional on permission, the full-rank information matrix establishes separate identifiability of \(q\) and \(r\). A model in which the permission state \(Z\) is itself learned can incorporate additional evidence from structural cues, known permitted expressions, and repeated histories of linguistic experience. Such observations provide complementary information about the grammatical status of the expression and its expected use.

For the simulation in Figure 4, we generated 4,000 datasets at each of \(N=32,64,128,256,512\), using \(q=0.35\) and \(r=0.60\). With opportunity labels the estimators are \(\hat{q}=N_{O}/N\) and \(\hat{r}=N_{Y}/N_{O}\), while the combined rate is estimated as \(N_{Y}/N\). The first and third estimators use all observations. The conditional choice estimator requires at least one observed opportunity (\(N_{O}>0\)). The simulation examines the precision of these estimators as labeled observations accumulate.

\section{S7. Shared continuations and generalization to unseen
combinations}\label{s7.-shared-continuations-and-generalization-to-unseen-combinations}

Shared distributional contexts provide cues to syntactic categories in child-directed speech \cite{si16,si17} and in artificial-language learning \cite{si01}. Distributional training also shapes generalization over argument structures \cite{si18}. Our task isolates one form of this inference: can the learner judge a word combination to be compatible using evidence from the words\textquotesingle{} separate predictive histories, when the combination itself has never occurred?

Each artificial grammar has three balanced categories, with four determiners, twelve nouns, and four continuation tokens per category. Nine nouns in each category occur in determiner frames, while the remaining three occur only after a neutral initial frame. Thus, for each determiner, all tested pairings with these nine neutral-frame nouns (three from each category) have zero training adjacency counts. The learner can nevertheless compare the continuation distributions associated with the two component words.

If a fraction \(\epsilon\) of continuation tokens is drawn independently of category, the category-specific component of each distribution has weight \(1-\epsilon\). Centering on the overall continuation distribution removes the common background, leaving category-linked differences of this size. The weighted product of two centered profiles therefore scales with \((1-\epsilon )^{2}\). For \(K\) balanced categories, the population difference between within- and between-category SCB scores is \(K(1-\epsilon )^{2}\). We examined exposure effects across thirty generated grammars. At 120 training sentences, SCB achieved AUC 0.945 and direct-pair memory AUC 0.500; cosine similarity and \(k\)-means clustering of continuation profiles gave similar discrimination to SCB. Permuting the continuations across utterances left overall token frequencies and determiner--noun exposure unchanged but reduced SCB AUC to 0.512.

We also manipulated the strength of the category-linked predictive signal during generation. Increasing \(\epsilon\) introduces more category-independent continuations; a separate post-generation permutation removes the associations of continuations with the words while preserving marginal counts. With 4,000 sentences, SCB AUC remained approximately 0.994 at \(\epsilon =0.8\), then fell to 0.502 at full noise (\(\epsilon =1\)), when the continuations no longer carry category information. The scrambled condition remained near chance across noise levels. Figure~\ref{fig:S4} compares this manipulation with the analytical separation margin, using six generated grammars with 8,000 sentences for the empirical margin estimates.

\setcounter{figure}{3}

\begin{figure}
\centering
\includegraphics[width=0.98\linewidth,height=\textheight,keepaspectratio,alt={Shared continuations and the discrimination of unseen pairs. A. AUC after 4,000 training sentences as continuation noise increases. The shared-context bridge approaches chance at full noise, while permuting the continuations produces near-chance discrimination throughout. B. Mean within- versus between-category score difference across six generated grammars at 8,000 sentences. Error bars show standard deviations across grammars, and the line gives the predicted margin K(1-\textbackslash epsilon)\^{}2.}]{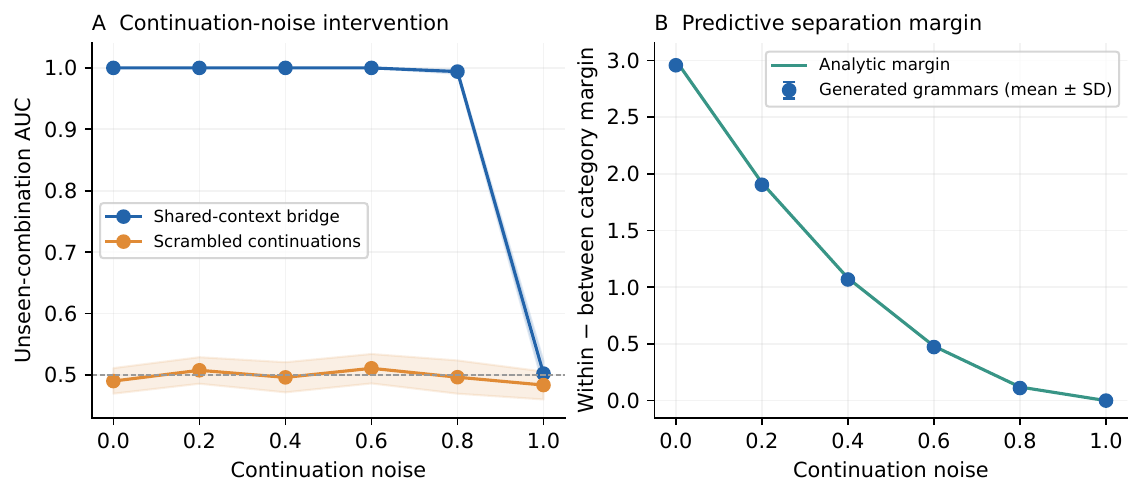}
\caption{\textbf{Category-linked continuations provide information about unseen pairings.} (A) AUC for the shared-context bridge at 4,000 training sentences as the proportion of category-independent continuations increases. Performance approaches chance when the category-linked signal is completely removed (noise = 1). The post-generation permutation control yields near-chance discrimination throughout. (B) Within-minus-between-category score difference across six generated grammars with 8,000 sentences each. Points and error bars represent the mean and standard deviation across grammars; the curve gives the analytical prediction \(K(1-\epsilon )^{2}\).}
\label{fig:S4}
\end{figure}

\setcounter{figure}{0}

The ability to recognize familiar sequences and the ability to infer the compatibility of novel combinations place different demands on a learning model. We examined these demands using an autoregressive gated recurrent unit (GRU) \cite{si19} with 40 hidden units, trained for 18 epochs on each of six artificial grammars. The comparison included a model with independently parameterized output weights and a tied-embedding variant, in which input embeddings and output weights are shared. Both models achieved perfect discrimination on directly observed regular vocabulary (AUC 1.0). For the unseen determiner--noun pairings, the corresponding mean AUCs were 0.556 and 0.484. The difference between performance on familiar and unseen combinations illustrates the particular computational demands of structural generalization from separate predictive histories.

The distinction between sequence recognition and structural generalization also arises in human statistical-learning experiments. In the dataset of Ordin et al. \cite{si20}, participants chose between trained words, phantom words, and nonwords. Phantom words are untrained sequences whose adjacent transitional statistics are matched to those of trained words. Across 28 participants and 2,016 choices, learners preferred trained words to phantom words on 55.80\% of trials (95\% participant-bootstrap interval [52.08\%, 59.38\%]). They preferred trained words to nonwords on 66.22\% of trials and phantom words to nonwords on 62.95\% of trials. These judgments reveal how experience with local sequential structure contributes to the recognition of linguistic units. Our artificial-grammar task examines a complementary form of inference, in which the separate predictive histories of two words provide evidence about their compatibility in a novel combination. Together, the two tasks illustrate different ways in which distributional experience can support linguistic knowledge.

\section{S8. How restriction judgments inform
production}\label{s8.-how-restriction-judgments-inform-production}

Work on statistical preemption has examined the restriction of adjective placement \cite{si21} and argument structures \cite{si22}. Samara et al.'s five experiments \cite{si02} provide a controlled comparison between constructions that compete to express the same meaning and constructions that serve different communicative functions. The archived sample sizes in the entrenchment and preemption conditions are, respectively, 43 versus 39, 40 versus 40, 40 versus 73, 40 versus 40, and 40 versus 45 in Experiments 1--5. The archive for Experiment 1 contains 82 participant records, and the published report describes a sample of 80 participants. To examine the sensitivity of the restriction effect to this difference, we evaluated all 3,321 possible subsets obtained by excluding two participants from the archive. Across these subsets, the estimated increase in restriction under preemption ranged from 0.892 to 1.176 rating points.

To ask whether knowledge expressed in a judgment task also informs a learner\textquotesingle s production, we linked judgments and productions within each experiment. A production was eligible when it involved a restricted training word, the scene called for the combination not attested for that word in training, and the response used one of the two marked constructions. The binary outcome indicates whether the learner produced the unattested restricted combination. Across 399 participants, there were 3,162 eligible productions. The reference model uses condition and general judgment ratings, and the extended model adds the ratings of unattested restricted combinations.

Some participants contribute more eligible responses than others, so we summarize the improvement in two ways. For participant \(i\), let \(n_{i}\) be their valid production count and \(\Delta _{i}\) their average held-out per-production log-score gain \cite{si23}. We compute

\[\Delta_{\mathrm{person}}=\frac1{N_p}\sum_i\Delta_i,\qquad \Delta_{\mathrm{production}}=\frac{\sum_i n_i\Delta_i}{\sum_i n_i}.\]

The first gives the same weight to each learner and the second gives the same weight to each production. Both use the same held-out predictions from binomial fits to \((k_{i},n_{i})\), where \(k_{i}\) counts restricted choices. The 95\% percentile bootstrap intervals resample learners together with their counts and predictions. Table~\ref{tab:S2} compares the average gains when either participants or entire experiments are withheld.

\begin{longtable}[]{@{}
  >{\raggedright\arraybackslash}p{(\linewidth - 8\tabcolsep) * \real{0.2500}}
  >{\raggedright\arraybackslash}p{(\linewidth - 8\tabcolsep) * \real{0.3088}}
  >{\raggedleft\arraybackslash}p{(\linewidth - 8\tabcolsep) * \real{0.1471}}
  >{\raggedleft\arraybackslash}p{(\linewidth - 8\tabcolsep) * \real{0.1471}}
  >{\raggedleft\arraybackslash}p{(\linewidth - 8\tabcolsep) * \real{0.1471}}@{}}
\caption{Gains from ratings of unattested restricted combinations when predicting production choices for held-out participants or experiments. Gains are in nats per production; intervals are 95\% participant-bootstrap intervals.\label{tab:S2}}\tabularnewline
\toprule\noalign{}
\begin{minipage}[b]{\linewidth}\raggedright
Held-out unit
\end{minipage} & \begin{minipage}[b]{\linewidth}\raggedright
Average
\end{minipage} & \begin{minipage}[b]{\linewidth}\raggedleft
Gain
\end{minipage} & \begin{minipage}[b]{\linewidth}\raggedleft
Lower
\end{minipage} & \begin{minipage}[b]{\linewidth}\raggedleft
Upper
\end{minipage} \\
\midrule\noalign{}
\endfirsthead
\toprule\noalign{}
\begin{minipage}[b]{\linewidth}\raggedright
Held-out unit
\end{minipage} & \begin{minipage}[b]{\linewidth}\raggedright
Average
\end{minipage} & \begin{minipage}[b]{\linewidth}\raggedleft
Gain
\end{minipage} & \begin{minipage}[b]{\linewidth}\raggedleft
Lower
\end{minipage} & \begin{minipage}[b]{\linewidth}\raggedleft
Upper
\end{minipage} \\
\midrule\noalign{}
\endhead
\bottomrule\noalign{}
\endlastfoot
Experiment & Participant-equal & 0.0322 & 0.0103 & 0.0550 \\
Experiment & Production-weighted & 0.0316 & 0.0099 & 0.0543 \\
Participant & Participant-equal & 0.0365 & 0.0140 & 0.0604 \\
Participant & Production-weighted & 0.0359 & 0.0135 & 0.0598 \\
\end{longtable}

The gains are similar under the two weighting schemes and positive in both holdout designs. To examine how consistent the effect is across studies, Table~\ref{tab:S3} gives the separate experiment contributions. The additional restricted ratings improve predictions in every experiment when entire experiments are held out. The largest gains occur in Experiments 4 and 5. In the participant-held-out analysis, the point estimate for Experiment 1 is slightly negative.

\begin{longtable}[]{@{}
  >{\raggedright\arraybackslash}p{(\linewidth - 4\tabcolsep) * \real{0.1972}}
  >{\raggedleft\arraybackslash}p{(\linewidth - 4\tabcolsep) * \real{0.3944}}
  >{\raggedleft\arraybackslash}p{(\linewidth - 4\tabcolsep) * \real{0.4085}}@{}}
\caption{Experiment-specific gains in production prediction from restricted-combination ratings, with entire experiments or individual participants held out (nats per production).\label{tab:S3}}\tabularnewline
\toprule\noalign{}
\begin{minipage}[b]{\linewidth}\raggedright
Experiment
\end{minipage} & \begin{minipage}[b]{\linewidth}\raggedleft
Experiment-held-out gain
\end{minipage} & \begin{minipage}[b]{\linewidth}\raggedleft
Participant-held-out gain
\end{minipage} \\
\midrule\noalign{}
\endfirsthead
\toprule\noalign{}
\begin{minipage}[b]{\linewidth}\raggedright
Experiment
\end{minipage} & \begin{minipage}[b]{\linewidth}\raggedleft
Experiment-held-out gain
\end{minipage} & \begin{minipage}[b]{\linewidth}\raggedleft
Participant-held-out gain
\end{minipage} \\
\midrule\noalign{}
\endhead
\bottomrule\noalign{}
\endlastfoot
E1 & 0.0030 & -0.0002 \\
E2 & 0.0048 & 0.0056 \\
E3 & 0.0254 & 0.0180 \\
E4 & 0.0961 & 0.1108 \\
E5 & 0.0402 & 0.0744 \\
\end{longtable}

Different kinds of judgments may carry different amounts of information about a learner's production choices. To examine this distinction, we compared the predictive contributions of restricted-combination ratings and alternating-item ratings. The latter contributed mean gains of 0.0020 nats per production when participants were held out and 0.0010 when entire experiments were held out, giving each learner equal weight. Table~\ref{tab:S4} compares these judgment measures on a common scale. Ratings of unattested restricted combinations provide the largest additional predictive gain, highlighting the contribution of construction-specific restriction knowledge to active production.

\begin{longtable}[]{@{}
  >{\raggedright\arraybackslash}p{(\linewidth - 6\tabcolsep) * \real{0.5000}}
  >{\raggedleft\arraybackslash}p{(\linewidth - 6\tabcolsep) * \real{0.1613}}
  >{\raggedleft\arraybackslash}p{(\linewidth - 6\tabcolsep) * \real{0.1774}}
  >{\raggedleft\arraybackslash}p{(\linewidth - 6\tabcolsep) * \real{0.1613}}@{}}
\caption{Participant-equal gains from different judgment measures in the participant-held-out production task (nats per production; 95\% participant-bootstrap intervals).\label{tab:S4}}\tabularnewline
\toprule\noalign{}
\begin{minipage}[b]{\linewidth}\raggedright
Added information
\end{minipage} & \begin{minipage}[b]{\linewidth}\raggedleft
Gain
\end{minipage} & \begin{minipage}[b]{\linewidth}\raggedleft
Lower
\end{minipage} & \begin{minipage}[b]{\linewidth}\raggedleft
Upper
\end{minipage} \\
\midrule\noalign{}
\endfirsthead
\toprule\noalign{}
\begin{minipage}[b]{\linewidth}\raggedright
Added information
\end{minipage} & \begin{minipage}[b]{\linewidth}\raggedleft
Gain
\end{minipage} & \begin{minipage}[b]{\linewidth}\raggedleft
Lower
\end{minipage} & \begin{minipage}[b]{\linewidth}\raggedleft
Upper
\end{minipage} \\
\midrule\noalign{}
\endhead
\bottomrule\noalign{}
\endlastfoot
General ratings over condition & 0.0133 & -0.0017 & 0.0301 \\
Restricted over general ratings & 0.0365 & 0.0140 & 0.0604 \\
Alternating over general ratings & 0.0020 & -0.0063 & 0.0105 \\
\end{longtable}

\begin{figure}
\centering
\includegraphics[width=0.92\linewidth,height=\textheight,keepaspectratio,alt={Item-class effects after accounting for condition-specific use of the rating scale. Points compare held-out ordinal log scores after adding novel and alternating item interactions to a model that already allows condition-specific thresholds and a change for unattested restricted items. Horizontal intervals are 95\% percentile intervals obtained by resampling participants within each experiment, then averaging experiment gains equally for the pooled estimate. The dashed line marks zero gain.}]{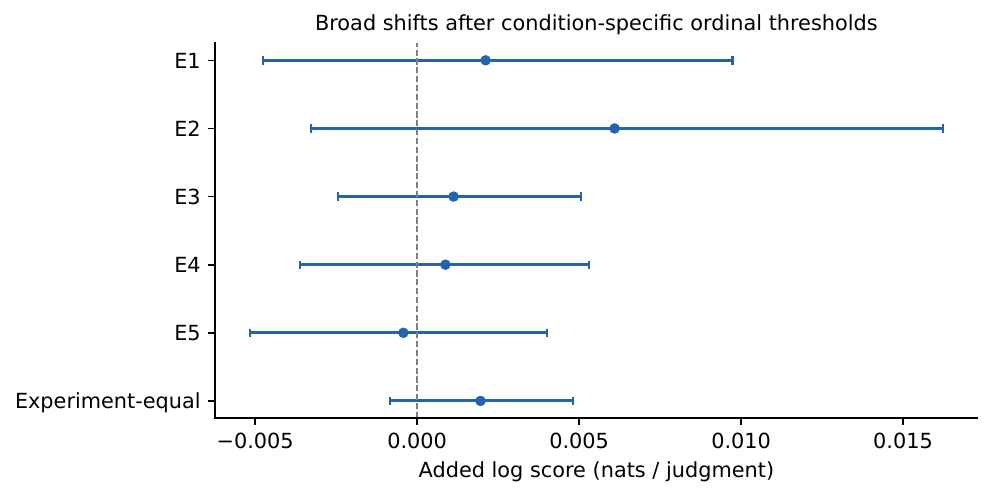}
\caption{\textbf{Additional item-class shifts after allowing for condition-specific rating thresholds.} Each point gives the held-out ordinal log-score gain from adding interactions for novel and alternating items to a reference model with condition-specific thresholds and an effect for unattested restricted items. Horizontal intervals are 95\% percentile bootstrap intervals, obtained by resampling participants within each experiment; the pooled estimate averages experiments equally. The dashed vertical line denotes zero additional predictive gain.}
\label{fig:S1}
\end{figure}

General response tendencies may also affect how learners judge different classes of items. In the raw scores, the difference between novel and attested items differs by training condition: the preemption-minus-entrenchment contrast is \(-1.006\) rating points (95\% bootstrap interval \([-1.194,-0.815]\)). We used an ordinal threshold model \cite{si24} to investigate whether broader item-class differences persist after allowing conditions to differ in their use of the rating scale. The reference model includes four rating thresholds for each training condition, item-class effects shared across conditions, and a condition effect specific to unattested restricted combinations. The expanded model additionally allows condition effects for novel and alternating items. The experiment-equal gain from the additional interactions is 0.0020 nats per judgment (95\% bootstrap interval \([-0.0008,0.0048]\)). Figure~\ref{fig:S1} and Table~\ref{tab:S5} report the experiment-level comparisons.

\begin{longtable}[]{@{}lrrr@{}}
\caption{Predictive gains from allowing additional condition effects for novel and alternating items in the ordinal judgment model. The reference model includes condition-specific rating thresholds and a targeted effect for unattested restricted combinations (nats per judgment; 95\% participant-bootstrap intervals within experiments).\label{tab:S5}}\tabularnewline
\toprule\noalign{}
Experiment & Gain & Lower & Upper \\
\midrule\noalign{}
\endfirsthead
\toprule\noalign{}
Experiment & Gain & Lower & Upper \\
\midrule\noalign{}
\endhead
\bottomrule\noalign{}
\endlastfoot
E1 & 0.0021 & -0.0048 & 0.0097 \\
E2 & 0.0061 & -0.0033 & 0.0162 \\
E3 & 0.0011 & -0.0024 & 0.0051 \\
E4 & 0.0009 & -0.0036 & 0.0053 \\
E5 & -0.0004 & -0.0052 & 0.0040 \\
Experiment-equal & 0.0020 & -0.0008 & 0.0048 \\
\end{longtable}

Once the model accounts for differences in rating-scale thresholds and the targeted restriction effect, further condition effects for novel and alternating items contribute relatively little predictive information. Together with the alternating-item comparison, this pattern locates the information shared between judgment and production primarily in the ratings associated with linguistic restriction. A joint model of the two tasks can investigate how this shared knowledge influences acceptability judgments and utterance choices through their respective response processes.

Training in Experiments 1--5 involved 192, 192, 384, 168, and 168 trials, with 64, 64, 128, 56, and 56 exposures to restricted items, respectively. Exposure count varied between experiments and remained fixed within each experiment. The context-refinement prediction concerns the distribution of a fixed amount of experience across communicative environments. An experimental test of this prediction would therefore vary the allocation of observations across contexts within a common training design, while holding total exposure constant.

\section{S9. Variation in article use across families and
corpora}\label{s9.-variation-in-article-use-across-families-and-corpora}

If learners encounter different usage rates in different settings, a model allowing those rates to vary should give better predictions than one that assumes a common rate for a word. We examine this possibility using article use in the gold annotations of UD-English-CHILDES \cite{si08}, which brings together CHILDES transcripts \cite{si25} under Universal Dependencies annotation \cite{si26}. We restrict the analysis to alphabetic common-noun lemmas longer than one character. The reference model first estimates a global lexical Beta prior from adult training speech and then estimates distributions conditional on grammatical number and dependency role, with shrinkage selected using training sources. The source-varying extension further averages over a local rate drawn around the uncertain reference mean. In each cross-validation fold, observations from the held-out family or corpus are excluded before estimating any of these distributions and their hyperparameters.

Grammatical number is obtained from the noun tags NN and NNS; rare anomalous tags are assigned to an unknown category. Adult speech contributes 11,030 singular and 1,982 plural noun tokens, and child speech contributes 12,293 singular and 1,841 plural tokens. The analysis covers eleven families in the Brown, Providence, Weist, Braunwald, Kuczaj, and Thomas corpora.

The first comparison asks whether source variation improves held-out predictions for the individual families. Table~\ref{tab:S6} reports gains in joint marginal log score per noun token. The token-weighted estimate gives each evaluated noun token equal weight, while the family-equal estimate averages the per-token gains for the eleven held-out families. For the corresponding 95\% percentile intervals, we resample families with their total log-score gains and token counts, or their per-token gains, respectively. Both summaries favor allowing source variation, with a larger estimated gain in child speech.

\begin{longtable}[]{@{}llrrr@{}}
\caption{Gains from allowing source-dependent rates in leave-one-family-out article-use prediction (nats per noun token; 95\% family-bootstrap intervals).\label{tab:S6}}\tabularnewline
\toprule\noalign{}
Speech & Average & Gain & Lower & Upper \\
\midrule\noalign{}
\endfirsthead
\toprule\noalign{}
Speech & Average & Gain & Lower & Upper \\
\midrule\noalign{}
\endhead
\bottomrule\noalign{}
\endlastfoot
Adult & Family-equal & 0.0123 & 0.0096 & 0.0150 \\
Adult & Token-weighted & 0.0123 & 0.0091 & 0.0155 \\
Child & Family-equal & 0.0162 & 0.0110 & 0.0216 \\
Child & Token-weighted & 0.0171 & 0.0107 & 0.0233 \\
\end{longtable}

Families within a recording collection may share collection-specific characteristics. To assess prediction across broader source boundaries, we also withheld one entire corpus at a time and re-estimated the models on the remaining collections. Table~\ref{tab:S7} shows positive gains under both averaging schemes. The uncertainty intervals are obtained by resampling entire corpora, reflecting the broader source boundaries examined in this comparison.

\begin{longtable}[]{@{}llrrr@{}}
\caption{Gains from allowing source-dependent usage rates when predicting article use in held-out recording corpora. Results are summarized by giving equal weight either to corpora or to noun tokens (nats per noun token; 95\% corpus-bootstrap intervals).\label{tab:S7}}\tabularnewline
\toprule\noalign{}
Speech & Average & Gain & Lower & Upper \\
\midrule\noalign{}
\endfirsthead
\toprule\noalign{}
Speech & Average & Gain & Lower & Upper \\
\midrule\noalign{}
\endhead
\bottomrule\noalign{}
\endlastfoot
Adult & Corpus-equal & 0.0129 & 0.0105 & 0.0153 \\
Adult & Token-weighted & 0.0123 & 0.0098 & 0.0154 \\
Child & Corpus-equal & 0.0168 & 0.0113 & 0.0229 \\
Child & Token-weighted & 0.0167 & 0.0109 & 0.0221 \\
\end{longtable}

The preceding comparisons concern prediction. We also wanted to establish whether source heterogeneity could be detected directly in the distribution of article occurrences. Within each word-by-number-by-role stratum, we used multivariate hypergeometric allocation to redistribute the fixed total number of articles across sources while preserving their respective noun counts. This conditional randomization provides a homogeneous-use reference that holds the lexical and grammatical margins fixed. We retained strata with at least two families contributing at least two noun tokens each, at least eight noun tokens in total, and a mix of tokens with and without articles. For stratum \(s\) and family \(f\), let \(n_{sf}\) be the retained noun count, \(y_{sf}\) the corresponding article count, and \(\widehat p_s=\sum_f y_{sf}/\sum_f n_{sf}\) the pooled article-use rate. The within-stratum dispersion is the sum of Pearson contributions from all retained family cells,
\[
D=\sum_s\sum_f\frac{(y_{sf}-n_{sf}\widehat p_s)^2}{n_{sf}\widehat p_s(1-\widehat p_s)}.
\]
We compared \(D\) with the distribution obtained by reallocating the articles 1,999 times with the same margins.

In adult speech, the comparison covers 188 strata and 805 source cells; the observed within-stratum dispersion is 1,003.55, compared with a null expectation of 654.73. In child speech, 210 strata and 854 source cells give an observed dispersion of 1,007.02 and a null expectation of 687.35. Both dispersion tail probabilities are \(p=0.0005\). Figure 6g shows the accompanying excess of source cells containing no articles. The observed variation concerns the effective usage rate \(\theta\), which combines the availability of semantic opportunities with the conditional selection of linguistic forms. These findings establish source-dependent variation as an empirical feature of natural language use. Examining the semantic, pragmatic, developmental, and speaker-related characteristics of communicative environments offers a way to understand how such variation arises.

\subsection{S9.1 Four-family comparison with lexical and grammatical
controls}\label{s9.1-four-family-comparison-with-lexical-and-grammatical-controls}

The leave-one-source-out comparisons in S9 use a reference that accounts for lexical identity and grammatical structure. We also examined how the apparent advantage of source variation changes as the reference model becomes more informative, using the same seven-family training and four-family test split throughout this comparison. The split comprises 12,514 adult training and development utterances, 4,567 held-out adult utterances, and 4,992 held-out child utterances. Evaluation on familiar nouns covers 2,150 adult tokens (238 lemmas) and 2,161 child tokens (233 lemmas). Under the lexical-rate reference for this split, the fitted global prior is \(\operatorname{Beta}(1.09,1.54)\) and the local concentration \(\kappa =5.86\). In the broader leave-one-source-out analyses above, these quantities are estimated separately within each training fold.

A reference based on one point estimate of the usage rate per word provides a simple starting point. Relative to this reference, the source-varying model improves the mean log score by 0.0424 nats per noun token in adult speech and 0.0250 in child speech. A more informative reference integrates uncertainty about the lexical rates; including dependency role further accounts for differences associated with grammatical function. These controls reduce the amount of variation that can be attributed to the additional source-specific component. Table~\ref{tab:S8} compares the source-varying model with these progressively more informative reference models.

\begin{longtable}[]{@{}
  >{\raggedright\arraybackslash}p{(\linewidth - 8\tabcolsep) * \real{0.1493}}
  >{\raggedright\arraybackslash}p{(\linewidth - 8\tabcolsep) * \real{0.4030}}
  >{\raggedleft\arraybackslash}p{(\linewidth - 8\tabcolsep) * \real{0.1493}}
  >{\raggedleft\arraybackslash}p{(\linewidth - 8\tabcolsep) * \real{0.1493}}
  >{\raggedleft\arraybackslash}p{(\linewidth - 8\tabcolsep) * \real{0.1493}}@{}}
\caption{Source-variation gains for the four-family test split under progressively more informative lexical and dependency-role reference models (nats per noun token; 95\% word-bootstrap intervals).\label{tab:S8}}\tabularnewline
\toprule\noalign{}
\begin{minipage}[b]{\linewidth}\raggedright
Speech
\end{minipage} & \begin{minipage}[b]{\linewidth}\raggedright
Control
\end{minipage} & \begin{minipage}[b]{\linewidth}\raggedleft
Gain
\end{minipage} & \begin{minipage}[b]{\linewidth}\raggedleft
Lower
\end{minipage} & \begin{minipage}[b]{\linewidth}\raggedleft
Upper
\end{minipage} \\
\midrule\noalign{}
\endfirsthead
\toprule\noalign{}
\begin{minipage}[b]{\linewidth}\raggedright
Speech
\end{minipage} & \begin{minipage}[b]{\linewidth}\raggedright
Control
\end{minipage} & \begin{minipage}[b]{\linewidth}\raggedleft
Gain
\end{minipage} & \begin{minipage}[b]{\linewidth}\raggedleft
Lower
\end{minipage} & \begin{minipage}[b]{\linewidth}\raggedleft
Upper
\end{minipage} \\
\midrule\noalign{}
\endhead
\bottomrule\noalign{}
\endlastfoot
Adult & Lexical uncertainty & 0.0162 & 0.0081 & 0.0246 \\
Child & Lexical uncertainty & 0.0123 & 0.0012 & 0.0250 \\
Adult & Lexical + dependency role & 0.0155 & 0.0068 & 0.0246 \\
Child & Lexical + dependency role & 0.0104 & 0.0014 & 0.0194 \\
\end{longtable}

Table~\ref{tab:S9} reports the family-specific gains against the lexical-uncertainty reference. Gains are positive for adult speech in each of the four held-out families; child-speech gains vary, with a negative estimate for Emma. The heterogeneity of these gains illustrates why evaluating performance across several recording sources is informative.

\begin{longtable}[]{@{}lrr@{}}
\caption{Family-specific source-variation gains for adult and child speech on the four-family test split, relative to the lexical-uncertainty reference (nats per noun token).\label{tab:S9}}\tabularnewline
\toprule\noalign{}
Held-out family & Adult gain & Child gain \\
\midrule\noalign{}
\endfirsthead
\toprule\noalign{}
Held-out family & Adult gain & Child gain \\
\midrule\noalign{}
\endhead
\bottomrule\noalign{}
\endlastfoot
Emma & 0.0157 & -0.0052 \\
Eve & 0.0166 & 0.0119 \\
Thomas & 0.0172 & 0.0246 \\
Violet & 0.0038 & 0.0054 \\
\end{longtable}

Finally, the reference model can account for grammatical number as well as dependency role. With this additional control, the source-varying model continues to yield positive average gains on the same four held-out families (Table~\ref{tab:S10}). The successive comparisons show how much source-related predictive information remains after accounting for increasingly specific lexical and grammatical properties.

\begin{longtable}[]{@{}lrrr@{}}
\caption{Source-variation gains for the four-family test split after accounting for lexical identity, grammatical number, and dependency role (nats per noun token; 95\% bootstrap intervals over held-out families).\label{tab:S10}}\tabularnewline
\toprule\noalign{}
Speech & Gain & Lower & Upper \\
\midrule\noalign{}
\endfirsthead
\toprule\noalign{}
Speech & Gain & Lower & Upper \\
\midrule\noalign{}
\endhead
\bottomrule\noalign{}
\endlastfoot
Adult & 0.0124 & 0.0102 & 0.0136 \\
Child & 0.0078 & 0.0027 & 0.0137 \\
\end{longtable}

\section{S10. Complete absences, temporal order, and transfer across words}\label{s10.-long-absences-temporal-order-and-transfer-across-words}

\subsection{S10.1 Predicting the probability of article-free subsets}\label{sec:S10.1}

The results in S9 show that a source-varying model improves the overall prediction of article use. Our theory also depends specifically on the probability of observing complete absence, which is sensitive to the lower tail of the distribution of usage rates. We therefore examined whether the model accurately predicts the probability that a collection of noun tokens from a word--source cell contains no articles. For a cell with \(N\) noun tokens, \(K\) of which have articles, the observed probability of selecting a random subset of \(n\) tokens with no articles is

\[P_0(n)=\frac{\binom{N-K}{n}}{\binom{N}{n}}.\]

\begin{figure}
\centering
\includegraphics[width=0.98\linewidth,height=\textheight,keepaspectratio,alt={Probability of observing a subset with zero articles. Adult and child panels compare observed random subsets of noun tokens with predictions from a lexical Beta model and a model that allows local rates to vary by source. At 16 tokens the source-varying model comes close to the adult mean and overestimates absence in child speech. Each value concerns a subset drawn from the recorded word-source cells.}]{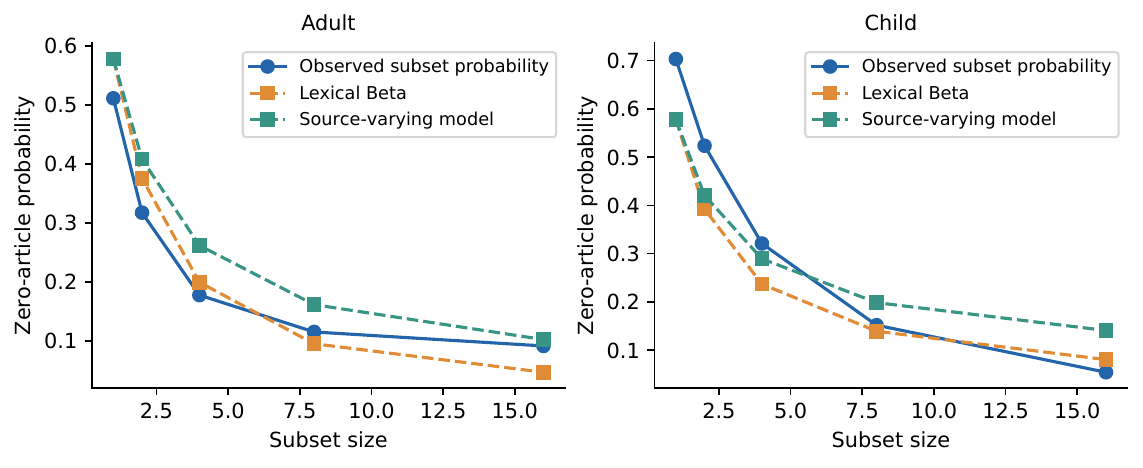}
\caption{\textbf{Probability that a randomly drawn subset of noun tokens contains no articles.} Adult and child panels compare the observed probabilities of selecting an article-free subset from a word--source cell with the corresponding predictions of the lexical Beta and source-varying models. Subset size is shown on the horizontal axis. At sixteen tokens, the source-varying model closely approximates the observed adult mean and predicts a higher probability of complete absence than observed in child speech. Each probability refers to a randomly drawn subset within a word--source cell.}
\label{fig:S2}
\end{figure}

This statistic concerns randomly selected subsets of noun occurrences, sampled without replacement from each word--source cell. Figure~\ref{fig:S2} compares these observed subset probabilities with predictions from the lexical Beta reference and the source-varying model. At \(n=16\), the observed mean, source-varying prediction, and lexical Beta prediction are respectively 0.0914, 0.1022, and 0.0468 in adult speech, and 0.0542, 0.1407, and 0.0803 in child speech. The comparison involves 24 eligible adult and 28 eligible child word--source cells.

The source-varying model closely approximates the observed mean probability of complete absence in adult speech. In child speech, its predicted probability is higher than the observed value, and the lexical Beta reference gives a closer estimate of the mean. Table~\ref{tab:S11} compares the two models using the Brier score \cite{si23}, with uncertainty intervals obtained by resampling words. The estimated Brier-score improvements have the same directional pattern as the mean predictions, with intervals spanning zero in both speaker groups. These comparisons highlight the importance of characterizing the low-usage regions of the distribution, particularly when the aim is to predict complete absence across relatively large collections of observations.

\begin{longtable}[]{@{}lrrr@{}}
\caption{Brier-score improvement for the source-varying model relative to the lexical Beta reference in predicting article-free subsets of sixteen noun tokens (95\% word-bootstrap intervals).\label{tab:S11}}\tabularnewline
\toprule\noalign{}
Speech & Brier-score gain & Lower & Upper \\
\midrule\noalign{}
\endfirsthead
\toprule\noalign{}
Speech & Brier-score gain & Lower & Upper \\
\midrule\noalign{}
\endhead
\bottomrule\noalign{}
\endlastfoot
Adult & 0.0095 & -0.0157 & 0.0428 \\
Child & -0.0207 & -0.0468 & 0.0013 \\
\end{longtable}

\subsection{S10.2 Does the recorded temporal order provide additional information?}\label{sec:S10.2}

The source-level and random-subset analyses characterize usage variation through the distribution of observations within recording sources. The temporal organization of those observations provides an additional dimension of linguistic experience. Nearby utterances may share a topic, speaker, or pragmatic state, while broader patterns of language use may develop with age. We therefore examined whether the recorded ordering of noun occurrences carries predictive information beyond their pooled frequencies. Tokens were ordered by source, age, sentence identifier, and token position. We used a renewal model in which a local usage rate persists across successive observations and is redrawn with probability \(h\) at each step, following the structure of constant-hazard change-point models \cite{si27}. Among the candidate renewal probabilities, \(h=0.05\) gave the best predictive performance on the training data.

\begin{figure}
\centering
\includegraphics[width=0.92\linewidth,height=\textheight,keepaspectratio,alt={Predictive information in the temporal order of article use. Bars give the score of the recorded sequence relative to the permutation mean for adult and child speech. Unrestricted permutations give the larger score difference. Permuting within six-month age bins preserves broad developmental composition and brings both differences closer to zero. Labels give permutation tail probabilities.}]{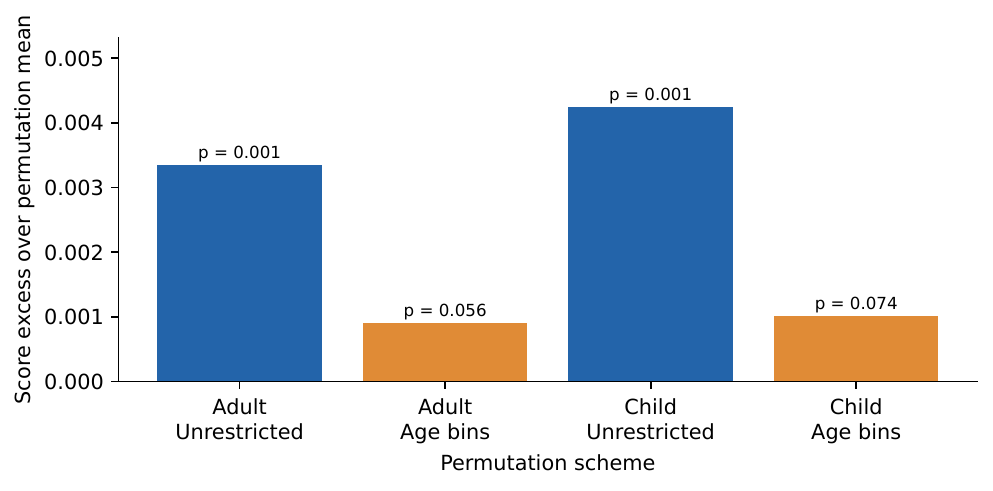}
\caption{\textbf{Additional predictive information in the recorded ordering of article-use observations.} Bars give the log-score excess of the recorded order over the mean of the permutation scores for adult and child speech. Unrestricted permutations disrupt both local ordering and developmental composition. Permutations within six-month age bins preserve broad age-related structure and yield smaller contrasts. Labels give permutation tail probabilities.}
\label{fig:S3}
\end{figure}

The recorded ordering provides higher predictive scores than random reorderings of the observations. Relative to 999 unrestricted permutations, the observed scores lie in the upper tail of the permutation distribution in both adult and child speech (\(p=0.001\)). Temporal ordering also carries information about developmental age. To examine the contribution of this broader structure, we repeated the analysis using permutations within six-month age intervals. Preserving the age composition produces smaller differences between the recorded and randomized orderings, with \(p=0.056\) for adult speech and \(p=0.074\) for child speech (Figure~\ref{fig:S3}). These comparisons highlight the contribution of age-related variation to the temporal organization of language use and provide a starting point for studying usage fluctuations over shorter discourse timescales.

\subsection{S10.3 A renewal model of persistent but changing local usage rates}\label{sec:S10.3}

The previous sections of the theory model a local usage rate that persists throughout a context. In natural speech, a context may end or change even when its boundary is not explicitly observed. We examine a mathematical extension in which each step can initiate a new persistent segment. The renewal rate \(h\) controls how often this happens. This provides a connection between the model with a single persistent rate and models in which new rates are repeatedly drawn.

Assume that every newly initiated segment draws a rate from \(\operatorname{Beta}(a,b)\). Its likelihood of \(k\) omissions, conditional on the segment persisting for those \(k\) observations, is \(m_{k}=B(a,b+k)/B(a,b)\). Let \(c=1-h\) be the probability that no reset occurs at a step. To obtain the likelihood \(L_{n}\) of \(n\) consecutive target omissions, we condition on the length of the first persistent segment. Either the segment covers all \(n\) observations, or it ends after some \(k<n\) and the remaining observations follow the same renewal mechanism. Thus

\[L_n = c^{n-1} m_n + \sum_{k=1}^{n-1} (1 - c) c^{k-1} m_k L_{n-k}, \quad c = 1 - h.\]

The first term is the contribution from a history without a reset; each term in the sum corresponds to a first reset after \(k\) observations. With \(L_{0}=1\) and \(m_{0}=1\), define

\[M(t)=\sum_{k=0}^{\infty}m_kt^k={}_2F_1(1,b;a+b;t).\]

The generating function of the omission likelihoods is

\[\sum_{n=0}^{\infty}L_nz^n=\frac{cM(cz)}{1-hM(cz)}.\]

Here \({}_2F_1\) denotes the Gaussian hypergeometric function. The generating function reveals two distinct mechanisms governing the long-run probability of complete absence. Their relative importance depends on how frequently the local usage rate is renewed.

For \(0<h<1\), the critical reset probability is
\[
h_c=
\begin{cases}
\dfrac{a-1}{a+b-1}, & a>1,\\[6pt]
0, & a\leq 1.
\end{cases}
\]
When \(a>1\), this threshold follows from the value of the generating function at its convergence boundary,
\[
M(1)=\frac{a+b-1}{a-1},
\]
at which the denominator vanishes when \(hM(1)=1\).

Below the threshold, the dominant singularity arises at the convergence boundary of \(M\). Long persistent segments with unusually low usage rates govern the asymptotic behavior, producing a power-law factor multiplying the exponential factor \(c^n\).

Above the threshold, the dominant singularity is a pole of the denominator at a positive argument inside the convergence boundary. Successive context renewals collectively determine the exponential decay rate of the omission likelihood. At a positive critical threshold, the pole meets the convergence boundary.

This transition has a mathematical parallel in the boundary-dominated and condensation-related regimes of the selection--mutation models developed by Kingman and by Dereich and M\"orters \cite{si06,si07}. In the present setting, it characterizes how the persistence and renewal of low-usage states determine the probability of prolonged linguistic absence. The resulting regimes provide theoretical predictions for models in which communicative states evolve over time.

\subsection{S10.4 Can usage rates be predicted for held-out words?}\label{sec:S10.4}

The SCB analysis in S7 examines whether shared predictive contexts provide evidence about the compatibility of unseen combinations. We now examine whether similar information can support quantitative predictions of form use. We represented 287 nouns with at least eight occurrences in the adult training data using 203 position-specific continuation features from the first and second following words, including an end-of-utterance marker. The profiles used gold-annotated adult speech from the training and development partitions. We retained features with at least five occurrences, divided their counts by the square root of their overall frequency, and normalized each noun's profile to unit length.

For each held-out noun, we averaged the \(\operatorname{Beta}(1,1)\)-smoothed article-use rates of fitting nouns, weighting each by the cosine similarity of its profile to that of the held-out noun raised to a power. We then shrank the estimate toward the token-weighted pooled rate of the fitting nouns, which also served as the prediction when the profile had no overlap with them. Five outer folds held out whole nouns. Within each outer training set, four inner folds selected the similarity power (1, 2, or 4) and the shrinkage weight (0, 0.25, or 0.5) by maximizing the log probability of article outcomes across tokens. For the corpus test partition, we evaluated the resulting rates with a Beta-binomial likelihood of concentration \(\kappa=4\) for each noun within each family, using the same concentration for the pooled-rate reference.

The original profiles yielded mean gains of \(-0.0003\) nats per token in adult speech and \(-0.0040\) in child speech relative to the pooled rate. We permuted the profiles among nouns within quartiles of adult training frequency 199 times and repeated model selection for each permutation. The mean permuted gains were \(0.0003\) nats per token for adult speech (central 95\% permutation range \([-0.0018,0.0022]\)) and \(-0.0016\) for child speech (\([-0.0038,-0.0003]\)). The observed gains fell below the permutation means in both groups (upper-tail \(p=0.845\) and \(p=0.985\), respectively).

We also represented nouns by the proportions of subject, object, root, and other occurrences in adult training speech. The mean gains over pooled rates were \(-0.0019\) nats per token in adult speech and \(-0.0032\) in child speech (Table~\ref{tab:S12}).

\begin{longtable}[]{@{}lrrr@{}}
\caption{Held-out article-use prediction gains using dependency-profile features (nats per token; 95\% word-bootstrap intervals).\label{tab:S12}}\tabularnewline
\toprule\noalign{}
Speech & Gain & Lower & Upper \\
\midrule\noalign{}
\endfirsthead
\toprule\noalign{}
Speech & Gain & Lower & Upper \\
\midrule\noalign{}
\endhead
\bottomrule\noalign{}
\endlastfoot
Adult & -0.0019 & -0.0071 & 0.0027 \\
Child & -0.0032 & -0.0076 & 0.0003 \\
\end{longtable}

Structural compatibility and actual usage involve different prediction problems. Shared continuations reveal category-linked structure when the linguistic environment supplies such information, as demonstrated in S7. The quantitative use of articles in natural speech also depends on the meanings, grammatical circumstances, and referring expressions associated with each noun. In the present analysis, pooled article-use rates provide similar or better predictions than estimates obtained from short continuations and dependency profiles. This pattern directs attention to richer forms of lexical, semantic, and competitive information, whose contributions we examine in S11.

\section{S11. Causative choices and judgments across five
languages}\label{s11.-causative-choices-and-judgments-across-five-languages}

Computational work on argument-structure acquisition has examined how learners combine distributional and semantic information about verbs \cite{si05}. To investigate the predictive contributions of these cues in different languages, we use a dataset of children\textquotesingle s judgments of alternative causative constructions. The task is particularly useful because the same broad causal meaning can be conveyed by several forms whose lexical and semantic distributions differ.

The archive of Ambridge et al. \cite{si09} contains 13,169 usable binary child judgments across 224 participant records. The numbers of judgments are 2,833 for English, 2,872 for Hebrew, 2,880 for Hindi, 2,665 for Japanese, and 1,919 for K'iche'. Each language contributes sixty verbs. The available lexical corpus contains three construction-specific counts and four rated semantic features for each verb. We ask whether models trained on these inputs can predict judgments of verbs whose behavioral data were withheld. In each fold, all child judgments for the test verbs are held out, but their corpus and semantic information remains available. This evaluation complements the original study\textquotesingle s unseen-corpus-verb task, in which corpus exposure to the test verbs is also withheld.

The softmax model learns to distribute its predicted output probability across competing forms. It receives sixty lexical identity indicators, one causative indicator, four semantic inputs, and a bias, which are connected directly to three outputs: less-transparent causatives, more-transparent causatives, and other forms. Let \(x_{i}\) denote an input row, \(y_{i}\) its one-hot target vector, \(w_{i}\) its sampled count, and \(W\) the weight matrix. Training minimizes the weighted cross-entropy with an L2 penalty:

\[J(W)=-\sum_iw_i\sum_ky_{ik}\log\operatorname{softmax}(x_iW)_k+0.5\|W\|_F^2.\]

With \(P\) denoting the matrix of predicted output probabilities and \(Y\) the target matrix, the analytical gradient is

\[\nabla J=X^\top[\operatorname{diag}(w)(P-Y)]+W.\]

Following the original implementation in \texttt{BinaryModeling.R}, we initialized the weights in the interval \([-0.5,0.5]\) and trained the network on forty-eight independently sampled corpora of 10,000 tokens each. Training used fifty BFGS iterations and the regularized objective defined above. The analytical gradient was evaluated against central finite differences. At prediction, the causative indicator was set to one, and the remaining inputs specified the identity of the verb and its semantic features. The predicted activation of each form was averaged across the trained networks.

Learning the distribution of linguistic forms from a corpus provides a basis for predicting how children evaluate those forms. We connect these two levels of prediction through a common behavioral model. For the softmax learner, the probability assigned to the judged form is transformed to a logit and combined with an indicator of that form. The additive and Beta-smoothed models use their corresponding frequency and semantic predictors, together with the form information specified by each model. All behavioral models use \(L_2\)-regularized logistic regression with an inverse regularization strength of \(C=1\) and are evaluated on the same five folds of verbs. Predictor scaling and the priors for frequency smoothing are estimated from the training verbs.

We first compare two ways of enriching the additive frequency-and-semantics reference: smoothing lexical construction rates and learning competitive form activations with softmax. The table below reports held-out log-score gains relative to the additive model. Each verb contributes its mean per-judgment gain, and the resulting means are averaged equally across verbs. The uncertainty intervals are obtained by resampling verbs. The clearest softmax gains over the additive reference occur in Japanese and K'iche', whereas the clearest benefit of Beta smoothing combined with semantics occurs in Hindi.

\Needspace{13\baselineskip}

\textbf{Held-out child-judgment log-score gains for softmax and Beta-smoothed semantic models relative to the additive frequency-and-semantics reference (nats per judgment; 95\% verb-bootstrap intervals).}

{\def\LTcaptype{none}
\begin{longtable}[]{@{}
  >{\raggedright\arraybackslash}p{(\linewidth - 4\tabcolsep) * \real{0.1519}}
  >{\raggedright\arraybackslash}p{(\linewidth - 4\tabcolsep) * \real{0.4177}}
  >{\raggedright\arraybackslash}p{(\linewidth - 4\tabcolsep) * \real{0.4304}}@{}}
\toprule\noalign{}
\begin{minipage}[b]{\linewidth}\raggedright
Language
\end{minipage} & \begin{minipage}[b]{\linewidth}\raggedright
Softmax vs.~additive
\end{minipage} & \begin{minipage}[b]{\linewidth}\raggedright
Smoothing vs.~additive
\end{minipage} \\
\midrule\noalign{}
\endhead
\bottomrule\noalign{}
\endlastfoot
English & 0.00132 {[}-0.00076, 0.00347{]} & -0.00007 {[}-0.00026,
0.00014{]} \\
Hebrew & -0.00048 {[}-0.00356, 0.00266{]} & -0.00019 {[}-0.00059,
0.00025{]} \\
Hindi & 0.00158 {[}-0.00164, 0.00512{]} & 0.00290 {[}0.00029,
0.00567{]} \\
Japanese & 0.00344 {[}0.00105, 0.00594{]} & 0.00014 {[}-0.00023,
0.00051{]} \\
K'iche' & 0.00242 {[}0.00002, 0.00495{]} & -0.00006 {[}-0.00013,
\(-4\times10^{-6}\){]} \\
\end{longtable}
}

The comparison with the additive reference helps isolate particular kinds of model improvement, but it is also informative to ask how well each individual cue predicts judgments. Tables~\ref{tab:S13} and \ref{tab:S14} report the average log scores of all seven models on the same held-out judgments. Higher scores indicate better predictions.

\begin{longtable}[]{@{}
  >{\raggedright\arraybackslash}p{(\linewidth - 8\tabcolsep) * \real{0.1967}}
  >{\raggedleft\arraybackslash}p{(\linewidth - 8\tabcolsep) * \real{0.1803}}
  >{\raggedleft\arraybackslash}p{(\linewidth - 8\tabcolsep) * \real{0.2131}}
  >{\raggedleft\arraybackslash}p{(\linewidth - 8\tabcolsep) * \real{0.2131}}
  >{\raggedleft\arraybackslash}p{(\linewidth - 8\tabcolsep) * \real{0.1967}}@{}}
\caption{Mean held-out judgment log scores for models using form, frequency, semantics, or additive frequency and semantic information. Higher scores indicate better prediction.\label{tab:S13}}\tabularnewline
\toprule\noalign{}
\begin{minipage}[b]{\linewidth}\raggedright
Language
\end{minipage} & \begin{minipage}[b]{\linewidth}\raggedleft
Form
\end{minipage} & \begin{minipage}[b]{\linewidth}\raggedleft
Frequency
\end{minipage} & \begin{minipage}[b]{\linewidth}\raggedleft
Semantics
\end{minipage} & \begin{minipage}[b]{\linewidth}\raggedleft
Additive
\end{minipage} \\
\midrule\noalign{}
\endfirsthead
\toprule\noalign{}
\begin{minipage}[b]{\linewidth}\raggedright
Language
\end{minipage} & \begin{minipage}[b]{\linewidth}\raggedleft
Form
\end{minipage} & \begin{minipage}[b]{\linewidth}\raggedleft
Frequency
\end{minipage} & \begin{minipage}[b]{\linewidth}\raggedleft
Semantics
\end{minipage} & \begin{minipage}[b]{\linewidth}\raggedleft
Additive
\end{minipage} \\
\midrule\noalign{}
\endhead
\bottomrule\noalign{}
\endlastfoot
English & -0.6895 & -0.6874 & -0.6871 & -0.6877 \\
Hebrew & -0.6945 & -0.6868 & -0.6951 & -0.6890 \\
Hindi & -0.6779 & -0.6689 & -0.6745 & -0.6685 \\
Japanese & -0.6923 & -0.6943 & -0.6932 & -0.6959 \\
K'iche' & -0.6398 & -0.6407 & -0.6414 & -0.6424 \\
\end{longtable}

\begin{longtable}[]{@{}
  >{\raggedright\arraybackslash}p{(\linewidth - 6\tabcolsep) * \real{0.1791}}
  >{\raggedleft\arraybackslash}p{(\linewidth - 6\tabcolsep) * \real{0.3284}}
  >{\raggedleft\arraybackslash}p{(\linewidth - 6\tabcolsep) * \real{0.3284}}
  >{\raggedleft\arraybackslash}p{(\linewidth - 6\tabcolsep) * \real{0.1642}}@{}}
\caption{Mean held-out judgment log scores for models based on Beta-smoothed frequency, Beta-smoothed frequency combined with semantics, and softmax corpus learning. Higher scores indicate better prediction.\label{tab:S14}}\tabularnewline
\toprule\noalign{}
\begin{minipage}[b]{\linewidth}\raggedright
Language
\end{minipage} & \begin{minipage}[b]{\linewidth}\raggedleft
Smoothed frequency
\end{minipage} & \begin{minipage}[b]{\linewidth}\raggedleft
Smoothed semantics
\end{minipage} & \begin{minipage}[b]{\linewidth}\raggedleft
Softmax
\end{minipage} \\
\midrule\noalign{}
\endfirsthead
\toprule\noalign{}
\begin{minipage}[b]{\linewidth}\raggedright
Language
\end{minipage} & \begin{minipage}[b]{\linewidth}\raggedleft
Smoothed frequency
\end{minipage} & \begin{minipage}[b]{\linewidth}\raggedleft
Smoothed semantics
\end{minipage} & \begin{minipage}[b]{\linewidth}\raggedleft
Softmax
\end{minipage} \\
\midrule\noalign{}
\endhead
\bottomrule\noalign{}
\endlastfoot
English & -0.6876 & -0.6878 & -0.6864 \\
Hebrew & -0.6869 & -0.6892 & -0.6895 \\
Hindi & -0.6651 & -0.6656 & -0.6670 \\
Japanese & -0.6942 & -0.6957 & -0.6924 \\
K'iche' & -0.6407 & -0.6425 & -0.6400 \\
\end{longtable}

The complete comparison reveals different sources of predictive information across languages. General preferences for particular construction forms are especially informative in Japanese and K'iche', where the form-only model achieves slightly higher mean scores than the softmax learner. In Hindi, smoothing lexical construction counts provides a clearer benefit. These patterns illustrate how construction form, lexical experience, semantic properties, and competition contribute to the prediction of children's acceptability judgments. The scene-level observation designs developed in S6 offer a complementary means of examining how these sources of knowledge relate to the availability of semantic opportunities and the conditional selection of linguistic forms.

\section{S12. Distinguishing the generating
mechanisms}\label{s12.-distinguishing-the-generating-mechanisms}

The three omission models in the main article make different predictions about how a fixed number of target observations contributes evidence when those observations are distributed across contexts. A practical question is whether the differences are sufficiently large to distinguish the models from finite experimental data. We examine this with a model-recovery simulation \cite{si28} in which the data-generating mechanism is known.

We generated observations under the known fixed-rate model, the persistent local-rate model, and the uncertain shared-mean model. All three use the same overall mean effective rate of 0.2 and are compared across eighteen observation-allocation designs. For each lexical item, the observation is binary: did its target occur at least once, or was it absent throughout the allocated observations? For each generating mechanism and sample size, we simulated 3,000 datasets. We then calculated the likelihood under each of the three candidate mechanisms, with their parameters fixed, and selected the model assigning the highest likelihood to the simulated data. This procedure measures how distinguishable the specified mechanisms are from the pattern of complete-absence outcomes alone.

\begin{longtable}[]{@{}lrr@{}}
\caption{Probability of correctly recovering the known generating mechanism from complete-absence outcomes across observation-allocation designs, at two numbers of independent items. Candidate model parameters are fixed.\label{tab:S15}}\tabularnewline
\toprule\noalign{}
Generating mechanism & 36 items & 450 items \\
\midrule\noalign{}
\endfirsthead
\toprule\noalign{}
Generating mechanism & 36 items & 450 items \\
\midrule\noalign{}
\endhead
\bottomrule\noalign{}
\endlastfoot
Fixed-rate & 0.824 & 0.986 \\
Persistent local-rate & 0.488 & 0.973 \\
Shared-mean & 0.924 & 1.000 \\
\end{longtable}

The ability to distinguish the generating mechanisms improves as the number of independent lexical items increases. With 36 items, the recovery probabilities are 0.824 for the fixed-rate model, 0.488 for the persistent local-rate model, and 0.924 for the shared-mean model. With 450 items, the corresponding probabilities increase to 0.986, 0.973, and 1.000. Across exposure-allocation designs, complete-absence outcomes therefore contain information about the structure of the underlying usage-rate uncertainty. These fixed-parameter recovery results provide a quantitative reference for developing experimental designs and extending the analysis to models with unknown usage-rate parameters and individual differences in learning.

The empirical analyses use the publicly available \href{https://osf.io/4jdvf/}{artificial-language archive of Samara et al.} \cite{si02}, the \href{https://osf.io/7f2dg/}{five-language causative-learning archive of Ambridge et al.} \cite{si09}, and the gold annotations in \href{https://aclanthology.org/2025.udw-1.6/}{UD-English-CHILDES} \cite{si08}. Analysis scripts and derived results are available from the corresponding author.

\section{References}\label{references}

\begingroup
\setstretch{1.02}
\setlength{\parskip}{3pt}
\interlinepenalty=10000

\endgroup